\documentclass{nature}

\usepackage{etoolbox}
\usepackage{sansmath}
\usepackage{microtype}
\usepackage{algorithm}
\usepackage{algpseudocode}
\usepackage{graphicx}
\usepackage{verbatim}
\usepackage{booktabs}
\usepackage[flushleft]{threeparttable}
\usepackage{multirow}
\usepackage{color,soul}
\usepackage{caption}
\usepackage{subcaption}
\usepackage[hyphens]{url}  
\usepackage{fancyhdr}
\usepackage[breaklinks=true,colorlinks,allcolors=blue]{hyperref}
\usepackage{tikz}
\usepackage{authblk}
\usepackage{textcomp}
\usepackage{tabularx}
\usepackage{dsfont}
\usepackage{amsfonts}
\usepackage{mathtools}
\usepackage[capitalise]{cleveref} 

\title{Active label cleaning for improved dataset quality\\under resource constraints}

\author[1,$\dagger$]{M\'elanie Bernhardt}
\author[1,$\dagger$]{Daniel C. Castro}
\author[1]{Ryutaro Tanno}
\author[1]{Anton Schwaighofer}
\author[1]{Kerem C. Tezcan}
\author[1]{\\Miguel Monteiro}
\author[1]{Shruthi Bannur}
\author[2]{Matthew P. Lungren}
\author[1]{Aditya Nori}
\author[1]{Ben Glocker}
\author[1]{\\Javier Alvarez-Valle}
\author[1,*]{Ozan Oktay}

\affil[1]{Health Intelligence, Microsoft Research Cambridge, Cambridge, CB1 2FB, UK}
\affil[2]{Department of Radiology, Stanford University, Palo Alto, CA 94304, USA \vspace{2mm}}

\affil[$\dagger$]{\small These authors contributed equally to this work.}
\affil[*]{\small Corresponding author: ozan.oktay@microsoft.com \vspace{3mm}}

\begin{document}

\twocolumn[{
\begin{@twocolumnfalse}
\maketitle

\maketitle

\begin{abstract}

Imperfections in data annotation, known as label noise, are detrimental to the training of machine learning models and have a confounding effect on the assessment of model performance. Nevertheless, employing experts to remove label noise by fully re-annotating large datasets is infeasible in resource-constrained settings, such as healthcare.

This work advocates for a data-driven approach to prioritising samples for re-annotation---which we term ``active label cleaning".
We propose to rank instances according to estimated label correctness and labelling difficulty of each sample, and introduce a simulation framework to evaluate relabelling efficacy.
Our experiments on natural images and on a new medical imaging benchmark show that cleaning noisy labels mitigates their negative impact on model training, evaluation, and selection.

Crucially, the proposed approach enables correcting labels up to 4\texttimes\ more effectively than typical random selection in realistic conditions, making better use of experts’ valuable time for improving dataset quality.

\end{abstract}

\end{@twocolumnfalse}}]


\section*{Introduction}

The success of supervised machine learning primarily relies on the availability of large datasets with high-quality annotations. However, in practice, labelling processes are prone to errors, almost inevitably leading to noisy datasets---as seen in ML benchmark datasets \cite{northcutt_nips2020_workshop}. Labelling errors can occur due to automated label extraction \cite{majkowska2020chest, NIHDataset}, ambiguities in input and output spaces \cite{beyer2020imagenet}, or human errors \cite{peterson2019human} (e.g.\ lack of expertise). At training time, incorrect labels hamper the generalisation of predictive models, as labelling errors may be memorised by the model resulting in undesired biases \cite{arpit2017closer, zhang2016understanding}.
At test time, mislabelled data can have detrimental effects on the validity of model evaluation, potentially leading to incorrect model selection for deployment as the true performance may not be faithfully reflected on noisy data. Label cleaning is therefore crucial to improve both model training and evaluation.

\begin{figure}
    \centering
    \footnotesize\sffamily\sansmath
    \includegraphics[width=\linewidth]{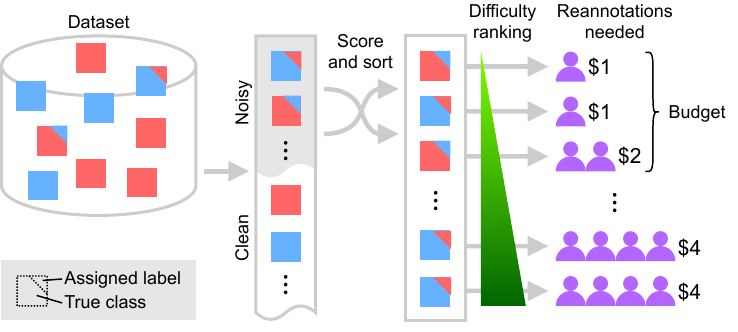}
    \caption{\textbf{Overview of the proposed active label cleaning.} A dataset with noisy labels is sorted to prioritise clearly mislabelled samples, maximising the number of corrected samples given a fixed relabelling budget.}
    \label{fig:overview}
\end{figure}

Relabelling a dataset involves a laborious manual reviewing process and in many cases the identification of individual labelling errors can be challenging. It is typically not feasible to review every sample in large datasets. Consider for example the NIH ChestXray dataset \cite{NIHDataset}, containing 112k chest radiographs depicting various diseases. Diagnostic labels were extracted from the radiology reports via an error-prone automated process \cite{OAKDENRAYNER2020106}. Later, a subset of images (4.5k) from this dataset were manually selected and their labels were reviewed by expert radiologists in an effort driven by Google Health \cite{majkowska2020chest}. Similarly, 30k randomly selected images from the same dataset were relabelled for the RSNA Kaggle challenge \cite{KaggleRSNA}. Such relabelling initiatives are extremely resource-intensive, particularly in the absence of a data-driven prioritisation strategy to help focusing on the subset of the data that most likely contains errors. 

\begin{figure}
    \centering
    \footnotesize\sffamily\sansmath
    \includegraphics[width=\linewidth]{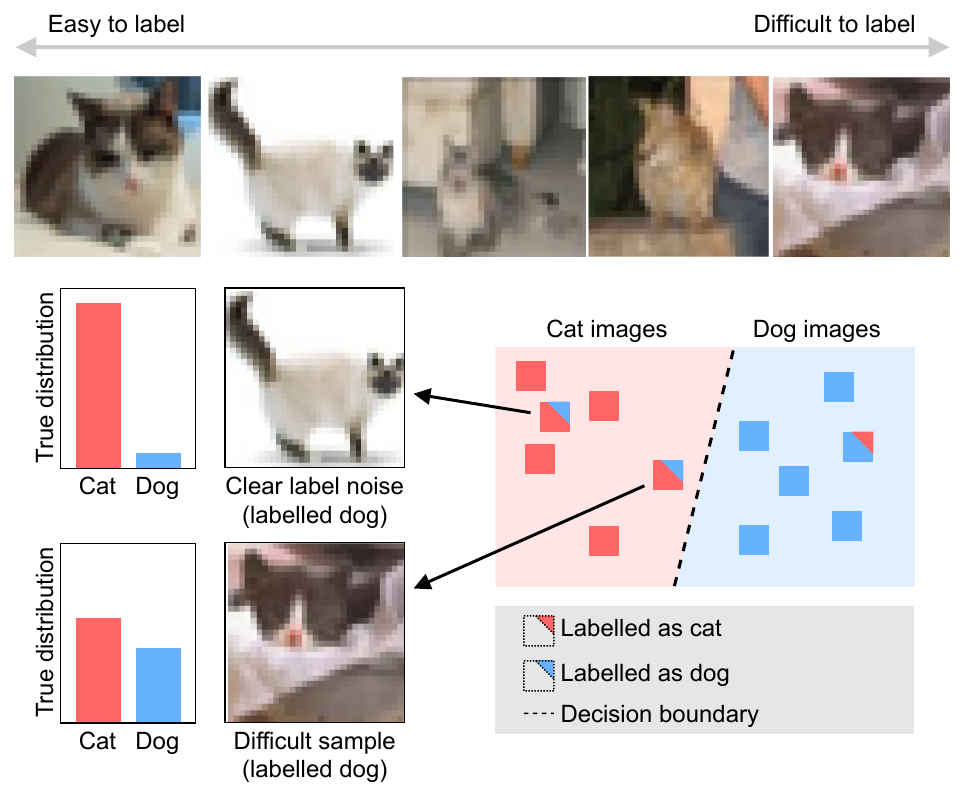}
    \caption{\textbf{Image labelling can become difficult due to ambiguity in input space \cite{battleday2020capturing}.} Top row shows the spectrum of ambiguity for cat images sampled from CIFAR10H dataset. The 2D plot illustrates different types of mislabelled samples: clear noise and difficult cases. We expect the former to be adjacent to semantically similar samples with a different label, and the latter to be closer to the optimal decision boundary.
    }
    \label{fig:sample_difficulty}
\end{figure}

Due to the practical constraints on the total number of re-annotations, samples often need to be prioritised to maximise the benefits of relabelling efforts (see \cref{fig:overview}), as the difficulty of reviewing labelling errors can vary across samples.
Some cases are easy to assess and correct, others may be inherently ambiguous even for expert annotators (\cref{fig:sample_difficulty}). For such difficult cases, several annotations (i.e.\ expert opinions) may be needed to form a ground-truth consensus \cite{krause2018grader, majkowska2020chest}, which comes with increasing relabelling ``cost". Hence, there is a need for relabelling strategies that consider both resource constraints and individual sample difficulty---especially in healthcare, where availability of experts is limited and variability of annotations is typically high due to the difficulty of the tasks \cite{gulshan2016development}.

While there are learning approaches designed specifically to handle label noise during training, we claim that these strategies can benefit from active labelling for two main reasons: First, clean evaluation labels are often unavailable in practice, in which case one cannot reliably determine whether any trained model is effective for a given real-world application. In that regard, active label collection can iteratively provide useful feedback to NRL approaches. Second, NRL approaches often cope with noise by inferring new labels \cite{li_dividemix} or disregarding samples \cite{yu19coteaching} that could otherwise be highly informative or even be correctly labelled.
However, models trained with these approaches can still learn biases from the noisy data, which may lead them to fail to identify incorrect labels, flag already correct ones, or even introduce additional label noise via self-confirmation.
Active label cleaning complements this perspective, aiming to correct potential biases by improving the quality of training dataset and preserving as many samples as possible. This is imperative in safety-critical domains such as healthcare, as model robustness must be validated on clean labels.

Prioritising samples for labelling also underpins the paradigm of active learning, whose goal is to select unlabelled samples that would be most beneficial for training in order to improve the performance of a predictive model on a downstream task. The key difference here for the proposed approach is that our goal is not only to improve model performance but also to maximise the quality of labels given limited resources, which makes it valuable for both training and evaluation of predictive models. In more detail, we demonstrate how active learning and NRL can play complementary roles in coping with label noise.

In this work, we begin by defining the active label cleaning setting in precise terms, along with the proposed relabelling priority score.
Using datasets of natural images and of chest radiographs, we then demonstrate experimentally the negative impacts of label noise on training and evaluating predictive models, and how cleaning the noisy labels can mitigate those effects.
Third, we show via simulations that the proposed active label cleaning framework can effectively prioritise samples to re-annotate under resource constraints, with substantial savings over naive random selection.
Fourth, we analyse how robust-learning \cite{han2018co} and self-supervision \cite{grill2020bootstrap} techniques can further improve label cleaning performance.
Lastly, we validate our choice of scoring function, which accounts for sample difficulty and noise level, comparing with an active learning baseline.

\section*{Results}
\subsection*{Active label cleaning.}
\label{sec:cleaning_scenario}
In this work, we introduce a sequential label cleaning procedure that maximises the number of corrected samples under a total resource budget $B \in \mathds{N}$:

\begin{equation}\label{eqn:relabelling_objective}
    \max \; \underbrace{\frac{1}{N} \sum_{i=1}^N \mathbf{1}[\hat{y}_i = y_i]}_\text{\clap{correctness of majority labels}} \quad
    \text{s.t.} \; \underbrace{\sum_{i=1}^N \lVert\hat{\mathbf{l}}_i\rVert_1 \leq B}_\text{budget constraint} \,,
\end{equation}
where we assume access to a dataset $\mathcal{D} = \{(\mathbf{x}_i, \hat{\mathbf{l}}_i)\}_{i=1}^N$, with the $i^\text{th}$ image $\mathbf{x}_i$, label counts vector ${\hat{\mathbf{l}}_i \in \mathds{N}^C}$ with $C$ classes, and corresponding majority label $\hat{y}_i  = \mathop{\mathrm{arg\,max}}_{c \in \{1,\dots,C\}} \hat{l}_i^c$.
All samples are assumed to initially contain at least one label ($\lVert\hat{\mathbf{l}}_i\rVert_1 = \sum_{c=1}^C \hat{l}_i^c \geq 1$), and some instances are mislabelled, i.e.\ their initial (majority) label $\hat{y}$ deviates from the true class $y$.
Unlike the traditional active learning objective, we aim to achieve a clean set of labels that can later be used not only for model training but also for benchmarking purposes. In that regard, our setting differs from active learning by allowing for re-annotation of already labelled instances as in `re-active' learning \cite{reactive_learning}. This is an important consideration as there is often a trade-off between label quality and downstream model performance \cite{ipeirotis2014repeated, reactive_learning}. Yet, conventional approaches \cite{ekambaram2016active, northcutt2019confident} may not be practically feasible with modern deep learning, as they often entail repeated retraining of models to measure the impact of each sample. 

The proposed framework (see \cref{algo:sampleselection}) determines relabelling priority based on predicted posteriors from a trained classification model, $p_{\boldsymbol{\theta}}(\hat{y} \vert \mathbf{x})$, parametrised by $\boldsymbol{\theta}$.
Label cleaning is performed over multiple iterations and at each iteration either a single or a batch of samples are relabelled.
Within an iteration, samples are first ranked according to predicted label correctness and ambiguity (i.e.\ annotation difficulty). Then, each prioritised sample is reviewed sequentially by different annotators until a majority is formed among all the collected labels $\hat{\mathbf{l}}_i$ (i.e.~$\hat{l}^{\hat{y}_i}_i > \hat{l}_i^c, \forall c \ne \hat{y}_i$, with $\sum_{c=1}^C \hat{l}_i^c > 1$). In the next iteration, the remaining samples are re-prioritised and the process repeats until the relabelling budget ($B$) is exhausted. Assuming that each annotation has a fixed cost, the more annotations a sample requires until a majority is reached, the more expensive its relabelling. Hence, to achieve the objective in \cref{eqn:relabelling_objective}, clearly mislabelled samples need to be prioritised over difficult cases, and correctly labelled samples should have the lowest relabelling priority.

To this end, we propose to rank available samples by the following scoring function $\Phi$:
\begin{equation}\label{eqn:scoring_fn}
\begin{split}
    \Phi(\mathbf{x}, \hat{\mathbf{l}}; \boldsymbol{\theta}) &=
        \underbrace{\mathrm{CE}(\hat{\mathbf{l}}, p_{\boldsymbol{\theta}})}%
        _\text{\clap{noisiness $\uparrow$}} \; - \;
        \underbrace{\mathrm{H}(p_{\boldsymbol{\theta}})}%
            _\text{\clap{ambiguity $\downarrow$}} \,.
\end{split}
\end{equation}
The first term, defined as the cross-entropy from the normalised label counts to the predicted posteriors,
\begin{equation}
    \mathrm{CE}(\hat{\mathbf{l}}, p_{\boldsymbol{\theta}}) =
        -\mathbb{E}_{\hat{\mathbf{l}}/ \lVert\hat{\mathbf{l}}\rVert_1}\left[\log p_{\boldsymbol{\theta}}(\hat{y} \vert \mathbf{x})\right] \,,
\end{equation}
corresponds to the estimated noisiness (i.e.\ negative log-likelihood) of the given labels.
Cross-entropy has been commonly used in NRL to detect mislabelled samples by ranking w.r.t.\ loss values \cite{han2018co,huang2019o2u,zhang_generalized_cross_entropy}, thus aiming to maximise the objective in \cref{eqn:relabelling_objective}.

On the other hand, obtaining a majority label vote requires different numbers of re-annotations for different images, depending on their difficulty. This is quantified by the entropy term $\mathrm{H}(p_{\boldsymbol{\theta}})$ defined over posteriors,
\begin{equation}
    \mathrm{H}(p_{\boldsymbol{\theta}}) =
        -\mathbb{E}_{p_{\boldsymbol{\theta}}(\hat{y} \vert \mathbf{x})}\left[\log p_{\boldsymbol{\theta}}(\hat{y} \vert \mathbf{x})\right] \,,
\end{equation}
which penalises ambiguous cases in the ranking.
This same quantity is employed as an estimate of aleatoric uncertainty (i.e.~data ambiguity) in active learning methods \cite{houlsby2011bald,kirsch2019batchbald}, which deprioritises ambiguous samples to address annotation budget constraints like ours in \cref{eqn:relabelling_objective}. This quantity can be better estimated by marginalising it over a distribution of $\boldsymbol{\theta}$ (Supplementary Fig.~6 and Table~2). Similar objectives have also been used in the contexts of semi-supervised learning \cite{grandvalet2004semi} and entropy regularisation \cite{pereyra2017regularizing}.
In Methods section, we present different options for the predictive model $p_{\boldsymbol{\theta}}$ to be used in computing these quantities.

Lastly, the proposed sequential framework allows for regular model updates during the cleaning process; as such, the selector model can be fine-tuned at any time using the corrected labels collected so far (see \cref{algo:sampleselection} \cref{alg:line:update})---which has shown to improve cleaning performance (Supplementary Table~3).

\subsection*{Datasets used in the experiments.}
To analyse label noise scenarios, we experiment with two imaging datasets, namely CIFAR10H and NoisyCXR, containing multiple annotations per data point, which helps us to model the true label distributions and associated labelling cost. In CIFAR10H \cite{battleday2020capturing, peterson2019human}, on average 51 manual annotations were collected for each image in the test set of CIFAR10 \cite{krizhevsky2009learning}, a collection of 10k natural images grouped in 10 classes. The experiments on CIFAR10H are intended to understand the challenges associated with different noise rates, noise models, and active relabelling scenarios. We also set up a new benchmark for label cleaning on medical images, NoisyCXR, comprising 26.6k chest radiographs and multiple labels from clinical datasets indicating the presence of pneumonia-like opacities \cite{KaggleRSNA, NIHDataset}. This serves as a more challenging imaging benchmark, where modelling difficulties are coupled with the challenges associated with noisy labels and sample prioritisation.

\begin{algorithm}[t]
    \footnotesize\sffamily
	\captionof{table}{Active label cleaning}
	\label{algo:sampleselection}
	\setlength{\tabcolsep}{1mm}
	\begin{tabularx}{\linewidth}{lX}
	\textbf{Given:}& $Y = \{\mathbf{l}_i\}_{i=1}^N$: True label distributions \\
    \textbf{Input:}& $\mathcal{D} = \{(\mathbf{x}_i, \hat{\mathbf{l}}_i)\}_{i=1}^N$: Dataset with noisy labels\\
                   & $B\in\mathds{N}$: Relabelling budget \\
                   & $b\in\mathds{N}$: Update frequency \\
    \end{tabularx}
	\begin{algorithmic}[1]
		\vspace{1mm}
        \State $\boldsymbol{\theta} \gets \Call{TrainRobustModel}{\mathcal{D}}$
  		\State $\mathcal{I}_\mathrm{avail} \gets \{1,\dots,N\}, \quad \mathcal{I}_\mathrm{cleaned} \gets \emptyset$
		\State count $\gets$ 0
		\While{count $<B$} \Comment{If budget remains}
    		\State $j \gets \mathop{\mathrm{arg\,max}}_{i \in \mathcal{I}_\mathrm{avail}} \Phi(\mathbf{x}_i, \hat{\mathbf{l}}_i; \boldsymbol{\theta})$ \Comment{Rank (\cref{eqn:scoring_fn})}
            \Repeat
    		    \State $\hat{\mathbf{l}}_j \gets \hat{\mathbf{l}}_j + \Call{Sample}{\mathbf{l}_j}$ \Comment{Acquire one-hot label}\label{alg:line:sample}
    		    \State count $\gets$ count + 1
    		\Until{majority formed in $\hat{\mathbf{l}}_j$}
     		\State $\mathcal{I}_\mathrm{avail} \gets \mathcal{I}_\mathrm{avail} \setminus \{j\}, \quad \mathcal{I}_\mathrm{cleaned} \gets \mathcal{I}_\mathrm{cleaned} \cup \{j\}$
     		\State $\mathcal{D} \gets \{(\mathbf{x}_i, \hat{\mathbf{l}}_i) : i\in\mathcal{I}_\mathrm{avail} \cup \mathcal{I}_\mathrm{cleaned}\}$
     		\If{count divisible by $b$}
     		    \State $\boldsymbol{\theta} \gets \Call{Update}{\boldsymbol{\theta}, \mathcal{D}}$ \Comment{Fine-tune model}\label{alg:line:update}
     	    \EndIf
		\EndWhile
		\State \Return{$\mathcal{D}$}
	\end{algorithmic}
\end{algorithm}

\subsection*{Label noise undermines both model training and evaluation.}
In particular, training with noisy labels is known to impair the performance of predictive models \cite{arpit2017closer, zhang2016understanding}, as the latter may be forced to memorise wrong labels instead of learning generalisable patterns. As an example, a vanilla convolutional neural network (CNN) classifier trained on CIFAR10H with clean labels achieves 73.6\% accuracy on a clean test set of 50k images. However, if the same model is trained on a version of this dataset with 30\% label noise, its test accuracy degrades to 64.1\%, a substantial drop of 8.5\%. Even a noise-robust classifier, initialised with self-supervision (i.e.\ pre-trained with images only), drops from 80.7\% to 78.7\% in accuracy when fine-tuned on the same noisy training labels as opposed to clean labels.

We also highlight the adverse implications of validation label noise on model evaluation. Its most evident impact is that true model performance can be underestimated. For instance, consider CIFAR10H with 40\% noise rate in both training and validation labels (instance-dependent noise \cite{xia2020part}; see Methods). A self-supervised classifier, as above, is estimated to be only 46.5\% accurate on these noisy validation labels, compared to 65.8\% on the true, clean labels. Additionally, model rankings based on noisy validation metrics may be unreliable, leading to misinformed model selection or design decisions. For example, a second classifier, trained on the same training set via co-teaching \cite{han2018co}, achieves a lower 45.9\% accuracy on the same noisy validation set, while its real accuracy (unobservable in practice) on clean labels is 69.9\%. In other words, the worst performing model (here, self-supervised) would have been chosen for deployment because of misleading validation metrics obtained on noisy labels.

In summary, label noise can negatively affect not only model building, but also validation. The latter is especially relevant in high-risk applications, e.g.\ for the regulation of models in healthcare settings. Our results in later sections demonstrate how active cleaning of noisy training and evaluation labels can help mitigate such issues.

\subsection*{Simulation of sequential relabelling.}
\label{sec:simulation}
A traditional way to evaluate a label cleaning algorithm is via its ability to separate noisy from clean labels (e.g.\ detection rate) \cite{ekambaram2016active, northcutt2019confident}. However, such an evaluation does not necessarily take into account the sequential nature of sample selection, label acquisition, and model updates. Further, it typically assumes all new labels are correct, neglecting the effect of sample ambiguity on how many annotations are required. Both are crucial factors in measuring the effectiveness of label cleaning efforts in terms of resource usage and final outcome.

To account for these factors, we propose a realistic simulation of the entire relabelling process that leverages the true label distribution of each sample. As shown in \cref{algo:sampleselection}, the simulation proceeds sequentially. Once a sample has been selected for relabelling, new labels are drawn from its true distribution (\cref{algo:sampleselection}, \cref{alg:line:sample}), which reflects the probabilities of real annotators assigning each class to a given sample. This enables more faithful modelling of the expected relabelling effort for each sample and of the likely label errors. Through this simulation, the performance of sample selection algorithms can be measured in terms of percentage of corrected labels in the dataset as a function of the number of re-annotations.

The relabelling simulation is used as a testbed to evaluate three different sample ranking algorithms: standard CNN (vanilla), co-teaching \cite{han2018co}, and self-supervised pre-training with fine-tuning \cite{grill2020bootstrap}. All approaches employed the same image encoder (ResNet-50), augmentations, and optimiser type (see supplement for implementation details). At each iteration, new labels are sampled from the true label distribution, and experiments are repeated with 5 independent random seeds. Performance of selection algorithms is compared to two baselines: the oracle and random selector, which determine respectively the upper and lower bounds. The random selector chooses the next sample to be annotated uniformly at random as done in previous studies \cite{KaggleRSNA, majkowska2020chest}. Conversely, the oracle simulates the ideal ranking method by having access to the true label distribution.
This enables the oracle to prioritise least difficult noisy samples first, hence maximising the number of corrected samples for a given resource constraint.
Because the oracle may still sample incorrect labels, we additionally include a ``minimal sampler" for reference, which systematically returns the ground-truth label without emulating the ambiguity of each sample.
The maximum relabelling budget ($B$) in the simulation is set to the expected number of re-annotations required by the oracle to clean all mislabelled samples.

\begin{figure*}[t]
    \footnotesize\sffamily\sansmath
    \centering
    \includegraphics[width=\linewidth]{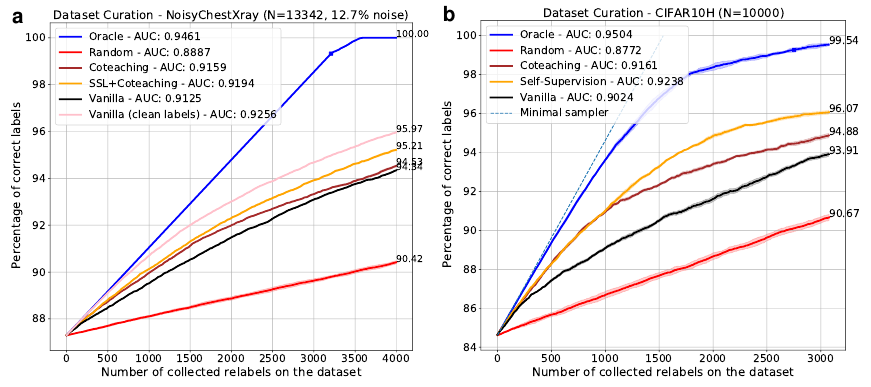}
    \caption{\textbf{Results of the label cleaning simulation on training datasets.} \textbf{a}~NoisyCXR\ ($\eta=12.7\%$); \textbf{b}~CIFAR10H ($\eta=15\%$). For a given number of collected labels ($x$-axis), a cost-efficient algorithm should maximise the number of samples that are now correctly labelled ($y$-axis). The correctness of acquired labels is measured in terms of accuracy. 
    The area-under-the-curve (AUC) is reported as a summary of cleaning efficiency of each selector across different relabelling budgets.
    The upper and lower bounds are set by oracle (blue) and random sampling (red) strategies. The pink curve (\textbf{a}) illustrates the practical ``model upper bound" of cleaning performance when the selector model is trained solely on clean labels, its performance being bound to the capacity of the model to fit the data. Shaded areas represent +/- standard deviation over 5 random seeds for relabelling.
    }
    \label{fig:simulation_results}
\end{figure*}

\begin{table}[t]
    \centering\footnotesize\sffamily\sansmath
	\setlength{\tabcolsep}{4pt}
	\begin{threeparttable}
	\caption{Classification accuracy (\%) before and after label cleaning.}
	\label{tab:classification_bald}
	\begin{tabular}{@{}clllp{12mm}p{12mm}@{}}
	    \toprule
	        & Selector  & Scoring & Classifier& Before cleaning & After cleaning \\
		\midrule
        (1) & Vanilla       & \cref{eqn:scoring_fn} & Vanilla   & 64.1  & 68.4 \\
        (2) & Vanilla       & BALD\cite{gal2017deep}& Vanilla   & 64.1  & 68.3 \\
        (3) & SSL           & \cref{eqn:scoring_fn} & Vanilla   & 64.1  & 70.9 \\
        (4) & SSL           & \cref{eqn:scoring_fn} & SSL       & 78.7  & 80.3 \\
        (5) &  Vanilla       & \cref{eqn:scoring_fn} &  SSL   & 78.7  & 79.4 \\
        (6) & Co-teaching   & \cref{eqn:scoring_fn} &Co-teaching& 66.5  & 68.8 \\
        (7) & --            & --        & ELR\cite{liu2020early}& 67.0  & -- \\
        \midrule
        (8) & \multicolumn{2}{c}{(Clean training)}  & Vanilla   & 73.6  & -- \\
        (9) & \multicolumn{2}{c}{(Clean training)}  & SSL       & 80.7  & -- \\
        \bottomrule
     \end{tabular}
     \scriptsize
     Models are evaluated on a clean test set ($N=50k$) before and after relabelling $32.7\%$ of samples in the training set (CIFAR10H, $N=5k$, $\eta=30\%$).
     \end{threeparttable}
\end{table}

\subsection*{Cleaning noisy training labels improves predictive accuracy.}
All label cleaning models are first trained on their respective training datasets (CIFAR10H and NoisyCXR), and subsequently used to rank samples within the same set for relabelling. In the rest of the manuscript, the term ``selector" refers to the models used for label cleaning. \Cref{tab:classification_bald} shows that all training methods benefit from newly acquired labels regardless of their underlying weight initialisation (e.g.\ self-supervised learning -- SSL) and noise-robust learning (NRL) strategy, as in ELR\cite{liu2020early}.
In that regard, label cleaning serves as a complementary approach to NRL by recycling data samples containing noisy labels. Additionally, the comparison between rows 3 and 7 show that acquiring a new set of labels can yield significantly better results than state-of-the-art NRL models trained on noisy labels, which performs closer to the upper-bound (row 8) where training is done with all true labels. Class imbalance, inherent noise model, and difficulty of samples are expected to introduce challenges to most NRL methods.

\subsection*{Label cleaning can be done in a resource-effective manner.}
\begin{figure*}[t]
    \centering
    \footnotesize\sffamily\sansmath
    \includegraphics[trim=1.25cm 0.8cm 1.0cm 1.5cm, clip, width=1.0\linewidth]{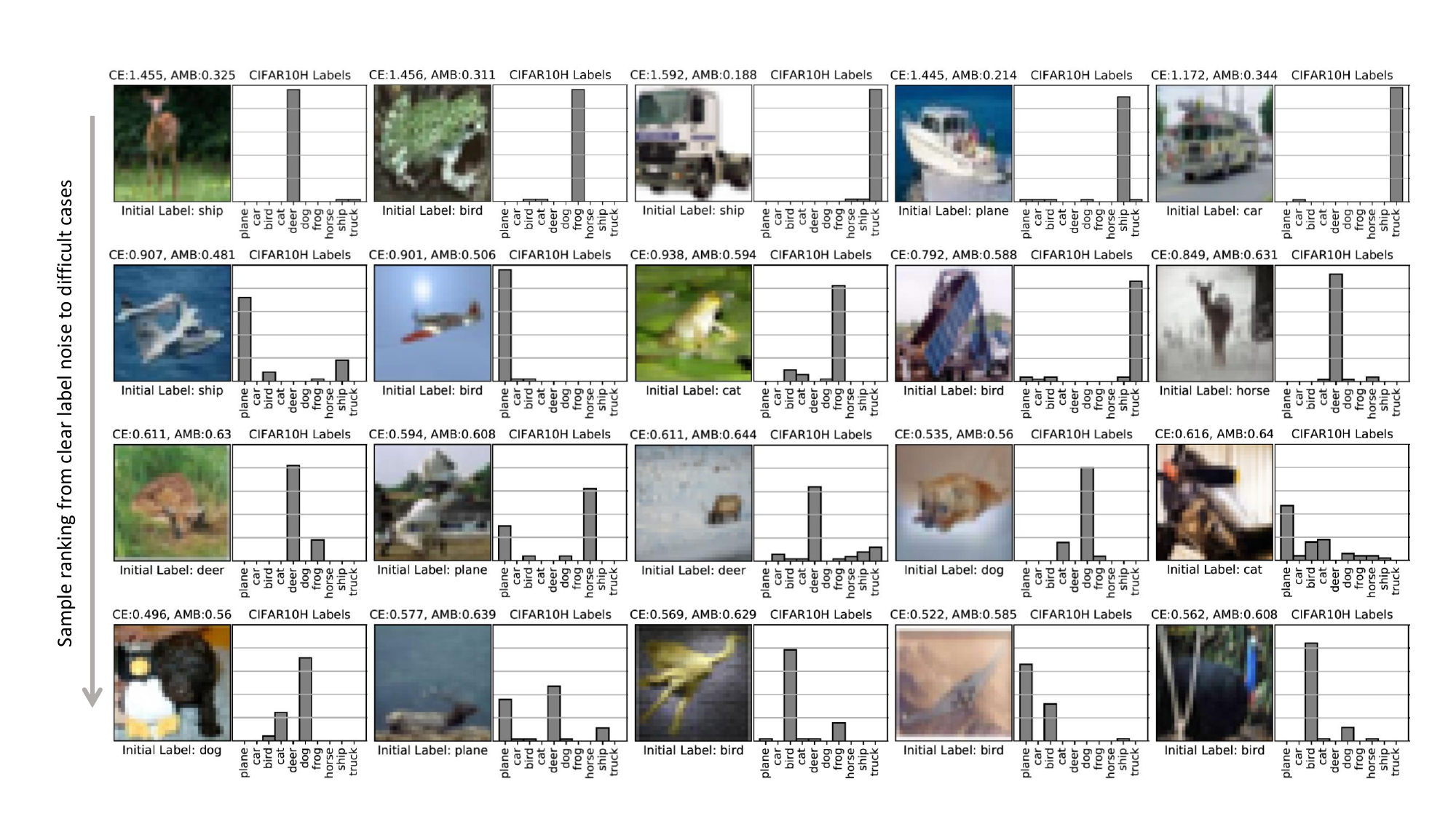}
    \caption{\textbf{Ranking of CIFAR10H samples (15\% initial noise rate) by the \textit{SSL-Linear} algorithm.} The top row illustrates a representative subset of images ranked at the top-10 percentile with the highest priority for relabelling. Similarly, the second and third rows correspond to 25--50 and 50--75 percentiles, respectively. At the bottom, ambiguous examples that fall into the bottom 10\% of the list ($N=2241$) are shown. Each example is shown together with its true label distribution to highlight the associated labelling difficulty. This can be compared against the label noisiness (cross-entropy; CE) and sample ambiguity (entropy; AMB) scores predicted by the algorithm (see \cref{eqn:scoring_fn}), shown above each image. As pointed out earlier, adjudication of samples provided at the bottom does require a large number of re-annotations to form a consensus. The authors in \cite{battleday2020capturing} explore the causes of ambiguity observed in these samples.}
    \label{fig:cifar10h_ranked_samples}
\end{figure*}

In \cref{tab:classification_bald}, we also observe that sample ranking methods (selectors) can influence the outcome of label cleaning efforts in terms of predictive accuracy (see rows 1~vs~3 and rows 4~vs~5). To further investigate the impact of selector types, we carried out label cleaning simulations (see \cref{fig:simulation_results}) on both CIFAR10H and NoisyCXR\ datasets for initial noise rates of $\eta=15\%$ and $12.7\%$, respectively. The methods are compared quantitatively regarding fraction of corrected labels (i.e.\ reduction in noise rate) for varying number of re-annotations. Additionally, the sample ranking is visually illustrated in \cref{fig:cifar10h_ranked_samples} along with the predicted label correctness and ambiguity scores.

\begin{figure}[t]
    \centering
    \footnotesize\sffamily\sansmath
    \includegraphics[width=\linewidth]{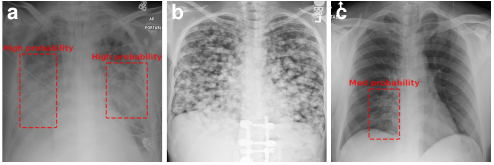}
    \caption{\textbf{Chest X-ray images selected from NoisyCXR\ dataset, which do not contain ``pneumonia" label in the NIH dataset \cite{NIHDataset}.} \textbf{a}~Correctly identified noise case with pneumonia-like opacities shown with bounding boxes. \textbf{b}~Wrongly flagged sample with a correct label; here the model confuses lung nodules with pneumonia-like opacities. \textbf{c}~A difficult case with subtle abnormality where radiologists indicated medium-confidence in their diagnosis as shown by the highlighted region (RSNA study \cite{KaggleRSNA}).}
    \label{fig:failure_cxr}
\end{figure}

It is apparent that, for a fixed improvement in number of corrected labels, all proposed selectors require much fewer re-annotations than random. For instance, the vanilla selector can reach 90\% correct labels with 3$\times$ (NoisyCXR) or 2.5$\times$ (CIFAR10H) fewer re-annotations than random. More specifically, we can see that noise-robust learning (here, co-teaching) benefits selection algorithms in prioritising noisy labels. Additionally, SSL yields further performance improvements by learning generic semantic representations without requiring labels, which in principle reduces chances of memorising noise. In that regard, it is a complementary feature to traditional noise-robust learning approaches as long as there is enough data for training. In particular, SSL pre-training yields improved noisy sample detection on NoisyCXR\ dataset in comparison to initialisation from scratch or pre-trained ImageNet weights. \Cref{fig:failure_cxr} exemplifies clear noisy and difficult cases for NoisyCXR.
We observed that the entropy term in \cref{eqn:scoring_fn} has had a smaller effect on the cleaning performance with NoisyCXR---which we hypothesise may be related to its binary labels---thus was not included in the simulation in \cref{fig:simulation_results}a.

The experiments are also repeated for larger noise rates (see Supplementary Figure~2) to identify the limits beyond which sample selection algorithms deteriorate towards random sampling. The results show similar performance as long as diagonal dominance \cite{chen2020robustness} holds on average in class confusion matrices (see \cref{fig:noise_model}), which is verified by experimenting with uniform and class-dependent noise models \cite{patrini2017making}.

\subsection*{Sample scoring function impacts cleaning performance.}
We study the influence of the self-entropy term ($\mathrm{H}(p_{\boldsymbol{\theta}})$) used in \cref{eqn:scoring_fn} in an ablation study. We see that selectors prioritising clear label noise cases yield higher numbers of corrected label noise, and this effect is further pronounced by including the entropy term in the scoring function (see Supplementary Figure~3).

Additionally, ranking samples for labelling is related to active learning, whereby one annotates unlabelled instances in order to maximise model performance metrics.
Typical active-learning methods attempt to identify the most informative examples \cite{toneva2018empirical}, relying on e.g.\ core-sets \cite{sener2018active} or high predictive uncertainty \cite{houlsby2011bald}.
A representative example is Bayesian active learning by disagreement (BALD) \cite{houlsby2011bald,gal2017deep}, which quantifies the mutual information between each sample's unknown label and the model parameters.
However, these criteria often prioritise under-represented parts of the data space, in particular samples close to the decision boundary. This may not necessarily coincide with samples with noisy labels, in particular clear noise cases (see \cref{fig:sample_difficulty}).
To analyse the suitability for label cleaning of such an approach, we repeat these experiments with the BALD score in replacement of the function in \cref{eqn:scoring_fn}, using the same relabelling budget.
Rows 1--2 in \cref{tab:classification_bald} show that classification accuracy gains on the test set are comparable between the two. However, the label quality improvement is significantly inferior for BALD (Supplementary Figure~3), as it favours samples with the highest disagreement---which may not correspond to label noise. Hence, whereas both scoring functions (BALD and \cref{eqn:scoring_fn}) yield similar model performance improvements, \cref{eqn:scoring_fn} offers the advantage of improving the quality of noisy datasets.

\subsection*{Cleaning evaluation labels mitigates confounding in model selection.}
\label{sec:model_ranking}
In noise-robust learning \cite{han2018co, li_dividemix, nguyen2020self, zhang_generalized_cross_entropy}, validation data is traditionally assumed to be perfectly labelled and can be relied on for model benchmarking and hyperparameter tuning. Furthermore, classification accuracy has been shown to be robust against noise \cite{chen2020robustness, lam2003evaluating} assuming that labelling errors and inputs are conditionally independent, $p(\hat{y} \vert y, \mathbf{x}) = p(\hat{y} \vert y)$.
While these assumptions may be valid in certain applications, they often do not hold for real-world datasets, wherein the labelling process clearly relies on examining the inputs.
Thus, models can learn biases from the training data that would correlate with labelling errors in the evaluation set (see \cref{fig:noise_model}).

To illustrate the pitfalls of evaluating models on noisy labels, we have conducted experiments on CIFAR10H using symmetric (\textit{SYM}) and instance-dependent noise (\textit{IDN}) models \cite{xia2020part} (see supplement for details).
In these experiments, diagonal dominance \cite{chen2020robustness} is preserved for all confusion matrices, and the statistical dependence between labelling errors, input, and true label are the same in both training and validation sets. Furthermore, all the available relabelling resources are used for cleaning the noisy labels in the evaluation set instead of training data to obtain an unbiased estimate of model ranking and performance. 
Though resources can be split to partially clean both training and evaluation sets without changes to the algorithm, we generally recommend prioritising the cleaning of evaluation data up to an acceptable level before applying any remaining budget to clean the training set.

Here, ResNet-50 models \cite{he2016deep} with different regularisation settings are trained with noisy training labels ($\vert\mathcal{D}_\mathrm{train}\vert = 10k$) and are later used to clean labels on evaluation set ($\vert\mathcal{D}_\mathrm{eval}\vert = 50k$). Labels for sets $\mathcal{D}_\mathrm{train}$ and $\mathcal{D}_\mathrm{eval}$ are determined for each experiment separately using \textit{SYM} and \textit{IDN} noise models with 40\% noise rate on average. The experimental results (\cref{tab:model_ranking_cifar}) show that standard performance metrics and model ranking strongly depend on the underlying conditions leading to labelling errors, which shows the need for clean validation sets in practical applications for model selection.

\begin{table}[t]
	\centering\footnotesize\sffamily\sansmath
	\begin{threeparttable}
	\caption{Classification accuracy under noisy evaluation labels.}
	\label{tab:model_ranking_cifar}
	\setlength{\tabcolsep}{3pt}
	\begin{tabular}{@{\extracolsep{1pt}}llcccc@{}}
	    \toprule
		& Model & True & Noisy (40\%) & Cleaned (10\%) & Cleaned (20\%)\\
		\midrule
        {\scriptsize\multirow{2}{*}{\rotatebox[origin=c]{90}{\textit{SYM}}}}
		& M1 	& 73.32 & 45.19 (.21) & 54.37 (.13) & 62.53 (.09) \\
		& M2 	& \textbf{80.16} & \textbf{48.93} (.23) & \textbf{58.45} (.10) & \textbf{67.42 (.10)} \\
		\midrule
        {\scriptsize\multirow{2}{*}{\rotatebox[origin=c]{90}{\textit{IDN}}}}
		& M1 	& \textbf{69.91} & 45.93 (.10)  & 49.97 (.06) & \textbf{55.87} (.06)\\
		& M2 	& 65.76     & \textbf{46.50} (.13) & 49.90 (.12) & 55.28 (.10)\\ 
		\bottomrule
	\end{tabular}
	\scriptsize
    Co-teaching (M1) and SSL (M2) models are compared on a noisy CIFAR10H validation set $\mathcal{D}_\mathrm{eval}$ over three runs using different label initialisations. Both approaches use ResNet-50 with different weight initialisation and regularisation. We compare classification accuracy on true, noisy, and cleaned labels. M2 is deliberately less regularised (i.e.\ weight decay) and is expected to perform worse on a more challenging \textit{IDN} noise model. For each validation set, the highest accuracy is highlighted in bold.
	\end{threeparttable}
\end{table}

To tackle these challenges, we applied the active relabelling procedure (SSL-Linear) on this noisy validation set by using the same corresponding training data ($\mathcal{D}_\mathrm{train}$). Through sample prioritisation and relabelling of 10\% of the entire set ($\mathcal{D}_\mathrm{eval}$), the bias in model selection can be alleviated as shown in \cref{tab:model_ranking_cifar}. At the end of the label cleaning procedure, the noise rates are reduced to 31.32\% (\textit{IDN}) and 30.25\% (\textit{SYM}); in other words, 86.8\% (\textit{IDN}) and 97.5\% (\textit{SYM}) of the selected images were initially mislabelled and then corrected. Here, \textit{IDN} model is observed to be more challenging for identifying noisy labels.

\section*{Discussion}\label{sec:discussion}
This work investigated the impact of label noise on model training and evaluation procedures by assessing its impact in terms of (I) predictive performance drop, (II) model evaluation results, and (III) model selection choices. As potential ways to mitigate this problem can be resource-demanding depending on the application area, we defined cost-effective relabelling strategies to improve the quality of datasets with noisy class labels. These solutions are benchmarked in a newly proposed simulation framework to quantify potential resource savings and improvement in downstream use-cases.

In particular, we highlight the importance of cleaning labels in a noisy evaluation set. We showed that neglecting this step may yield misleading performance metrics and model rankings that do not generalise to the test environment. This can, in turn, lead to overoptimistic design decisions with negative consequences in high-stakes applications such as healthcare.

\begin{figure}[t]
    \centering
    \footnotesize\sffamily\sansmath
    \includegraphics[width=\linewidth]{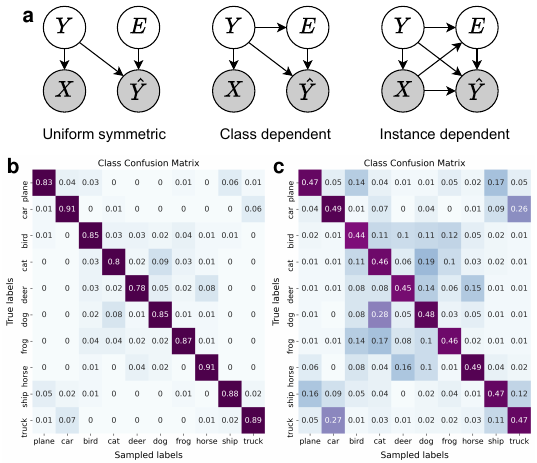}
    \caption{
    \textbf{Understanding label noise patterns.}
    \textbf{a} Different label noise models used in robust learning. The statistical dependence between input image ($X$), true label ($Y$), observed label ($\hat{Y}$), and error occurrence ($E$) is shown with arrows (adapted from Fr\'enay et al.\cite{frenay2014classification}).
    \textbf{b--c} CIFAR10H class confusion matrices (temperature $\tau=2$) for all samples (\textbf{b}) and difficult samples only (\textbf{c}).}
    \label{fig:noise_model}
\end{figure}

One of our main findings is that the patterns of label error in the data (i.e., structural assumptions about label errors shown in \cref{fig:noise_model}) can have as large an impact on the efficacy of label cleaning and robust-learning methods as the average noise rates, as evidenced by the results obtained on both training and validation sets. We therefore recommend carefully considering the underlying mechanisms of label noise when attempting to compare possible solutions.

Note that, when cleaning the test set, there may be concerns about introducing dependency between the training and test sets. To avoid this, selector models utilised for sample prioritisation should ideally (i)~be trained solely on the test set and (ii)~not be used for classification and evaluation purposes.
Moreover, it is worth noting that modelling biases could be reflected in the ranking of the samples.
Such biases may be mitigated by employing an ensemble of models with different formulations and inductive biases for posterior estimation in \cref{eqn:scoring_fn}, as the framework makes no assumptions about the family of functions that can be used for label cleaning.

The results also suggest that even robust-learning approaches may not fully recover predictive performance under high noise rates. In such cases, SSL pre-training is experimentally shown to be a reliable alternative, outperforming noise-robust models trained from scratch, even more so with the increasing availability of unlabelled datasets.
Lastly, we show that acquiring new labels can complement noise-robust learning by recycling data samples even if their labels are noisy, and can also handle biased labels. Thus, the two domains can be combined to obtain not only a better model, but also clean data labels for downstream applications.

A limitation of data-driven approaches for handling label noise is that they may still be able to learn from noise patterns in the data when label errors occur in a consistent manner. As such, some mislabellings may remain undetected.
However, it is worth noting that our approach does not flip already correct labels (assuming that manual labellers provide i.i.d.\ samples from the true data distribution). From this perspective, the proposed algorithm will converge towards the true label distribution (with a sufficient number of labels), addressing label inconsistencies detectable by the selector model while optimising for the objective given in \cref{eqn:relabelling_objective}.
Under extreme conditions where the bounded noise rate assumption may not hold (i.e.\ where on average there are more incorrect labels than correct in a given dataset), random selection can become preferable over data-driven approaches \cite{cheng2020learning}. 
However, in the case of bounded noise rate---as in most real-world applications---the active learning component of the proposed framework can potentially address such consistent noise patterns in labels\cite{cheng2020learning}.
Indeed, the proposed active approach enables establishing a distilled set of expert labels to tackle this challenge, instead of solely relying on self-distillation or hallucination of the true labels as in NRL methods\cite{cheng2020learning}.
To further extend our methodology along these lines, one could treat the newly acquired labels as expert-distilled samples, and rely more heavily on them for posterior updates in the proposed iterative framework.
Lastly, recall that the proposed label cleaning procedure is a human-in-the-loop system. Therefore, from a practical point-of-view, the process can be monitored and intervened upon whenever assumptions may be violated or if there is a concern around mislabelling biases in the dataset.

Although the present study focused on imaging, the proposed methodology is not limited to this data modality, and empirical validation with other input types is left for future work. It will also be valuable to explore, for example, having the option to also annotate unlabelled samples, or actively choosing the next annotator to label a selected instance. Such extensions to active cleaning will significantly broaden its application scope, enabling more reliable deployment of machine learning systems in resource-constrained settings.

Here, we assume that the majority vote was representative of the ground truth. However, in some cases majority vote can become suboptimal when annotators have different levels of experience in labelling data samples\cite{krause2018grader}. For these circumstances, future work could explore sample selection objectives and label assignment taking into account the expertise of each annotator. Similarly, multi-label fusion techniques \cite{khetan2017learning, rodrigues2013learning, tanno2019learning, warfield2004simultaneous} can be used within the proposed label cleaning procedure to restore true label distribution by modelling labelling process and aggregating multiple noisy annotations. Such approaches critically rely on the availability of multiple labels for each sample---which can be realised towards the end of relabelling efforts.

\footnotesize
\section*{Methods}
\subsection*{CIFAR10H dataset.}
\label{sec:cifar10h}

CIFAR10H \cite{battleday2020capturing, peterson2019human} is an image dataset comprising the test set of CIFAR10 (10k images) and multiple labels collected from crowdsourcing. Each image on average contains 51.1 labels distributed over 10 class categories. In our study, the true label distributions were calculated by normalising the histogram of these labels on a per image basis. In the experiments, we assume that our initial dataset contains labelling errors. These partially noisy labels were obtained by sampling the initial label by taking into account the label distribution for each sample. The noise rate in experiments is controlled by applying temperature scaling \cite{Guo2017} ($\tau$) to each distribution, which preserves the statistical dependence between input images and their corresponding labels. The higher the temperature, the closer to a uniform distribution the label distribution becomes, hence the higher the probability of sampling a noisy label for a given sample. 

Samples in this dataset are not all equally difficult to label or classify \cite{battleday2020capturing}. For instance, for some images, all annotators agreed on the same label, for some others collected labels were more diverse as the image was harder to identify. Hence, we defined sample difficulty as a function of the normalised Shannon entropy of its target distribution, $\sum_{c=1}^C p_c \log_C p_c$. If the entropy associated to this distribution was higher than a certain threshold (0.3 in our experiments), the sample was classified as ``difficult". The entropy of the target distribution can also be related to the expected number of relabellings required for this sample (see problem definition in Results). In \cref{fig:noise_model}, we show the average class confusion matrices over the CIFAR10H dataset for $\tau=2$. In \cref{fig:noise_model}, the confusion matrix of only the difficult samples is shown to highlight the classes that are confused the most in a standard labelling process.

\begin{table}
    \centering\footnotesize\sffamily\sansmath
    \begin{threeparttable}
    \caption{Correspondence of chest radiography labels from the RSNA challenge \cite{KaggleRSNA}, NIH \cite{NIHDataset}, and NoisyCXR\ datasets.}
    \label{table:xraylabels}
    \begin{tabular}{ @{\hspace{1em}} p{34mm} r r @{} } 
    \toprule
    & \multicolumn{2}{c}{RSNA labels \cite{KaggleRSNA}} \\
    \cmidrule{2-3}
    & Pneumonia-like & No pneumonia-like \\
    &  opacity & opacity \\
    \midrule
    \hspace{-1em} NIH labels \cite{NIHDataset}: \\
    Pneumonia & 367 & \textbf{441} \\ 
    Consolidation/infiltration (not pneumonia) & \textit{3,988} & \textit{8,551} \\ 
    Other diseases only & \textbf{1,101} & 5,567 \\
    No finding & \textbf{556} & 6,113 \\
    \midrule
    \hspace{-1em} NoisyCXR\ labels: \\
    Pneumonia-like opacity & 3,956 & \textbf{1,296} \\
    No pneumonia-like opacity & \textbf{2,056} & 19,376 \\
    \midrule
    \hspace{-1em} Total & 6,012 & 20,672 \\
    \bottomrule
    \end{tabular}

    \scriptsize
    The disagreements between two label sets are highlighted in bold font. Fields in italic indicate an unclear agreement between both sets of labels.
    NoisyCXR\ labels are collected from the original NIH labels where possible, and the remainder are uniformly sampled (10\%) from the ``Consolidation/infiltration" category to increase the noise rate further in the experiments.
    \end{threeparttable}
\end{table}

\subsection*{NoisyCXR: a medical benchmark dataset for label cleaning.}
\label{sec:chestxray}
In the experiments, we used a subset of the medical image dataset released by the NIH \cite{NIHDataset}. The original dataset contains 112k chest radiographs with labels automatically extracted from medical records using a natural language processing algorithm. In the original data release, labels are grouped into 14 classes (non-mutually exclusive) indicating the presence of various diseases or no finding at all. However, there have been studies \cite{OAKDENRAYNER2020106} showing errors in these auto-extracted image labels. Later, the RSNA released the Pneumonia Detection Challenge \cite{KaggleRSNA} aiming to detect lung opacities indicative of pneumonia. The challenge dataset is comprised of 30k images randomly sampled from the NIH ChestXray dataset mentioned above. To adjudicate the original labels released by the NIH, a board of radiologists reviewed and re-assessed each image on this dataset and delimited the presence of lung opacities associated to pneumonia. Samples are hence classified into two categories: ``pneumonia-like lung opacity" and ``no pneumonia-like lung opacity". Here we aim to utilise a realistically noisy version of this RSNA dataset encountered in real-world application scenarios. To this end, the original NIH labels are leveraged to construct a noisy version of the true labels released by the RSNA. In more detail, we analysed the data in four categories (see rows in \cref{table:xraylabels}) based on their original NIH labels and the class taxonomy \cite{irvin2019chexpert} in order to map this original label to a binary label indicating presence or absence of pneumonia-like opacities:
\begin{itemize}
    \itemsep0em 
    \item Cases where the NIH label contains ``Pneumonia". All these samples were assigned to the ``pneumonia-like opacity" category in NoisyCXR. 
    \item Cases associated to opacities potentially linked to pneumonia: NIH label contains ``consolidation" or ``infiltration" (but not ``pneumonia"). For these cases, it is unclear whether the original NIH label should be mapped to ``pneumonia-like opacity" or not \cite{irvin2019chexpert,oakdenrayner2017thoughts}. Hence, for these cases, in NoisyCXR, we attributed the correct (final RSNA) label to 90\% of the cases and flipped the label of the remaining 10\%. 
    \item Cases tagged as ``no finding", all assigned to the ``no pneumonia-like opacity" category. 
    \item Cases linked to other pathology (i.e., original label only contains pathology unrelated to pneumonia e.g.\ pleural effusion, nodules, fluid \cite{NIHDataset}), all assigned to the ``no pneumonia-like opacity" category. 
\end{itemize}

The obtained binary mapping hence gives us a set of noisy labels for the ``pneumonia-like opacity" classification task whereas the labels released as part of the challenge are considered as the correct labels for the data cleaning experiment. The final noisy dataset contains 26,684 images with a noise rate of 12.6\% (see \cref{table:xraylabels}). As in the case of CIFAR10H, the uncertainty in the final labels permits to separate ambiguous from easy cases. In particular, using the confidence associated to each pneumonia-like lung opacity bounding box \cite{kagglecxrboxes} we distinguished cases as follows:
\begin{itemize}
    \item If the final label was ``Pneumonia-like opacity" but all bounding boxes were of low confidence, we considered the case as difficult and define the label distribution for relabel sampling to $\mathbb{P}_\mathrm{label}(\text{Pneumonia opacity}) = 0.66$. If at least one of the boxes had a medium or high confidence score, the case is considered easy and we set $\mathbb{P}_\mathrm{label}(\text{Pneumonia opacity}) = 1$.
    \item If the final label was ``No Pneumonia-like opacity" but some readers delineated opacities, the sample is considered ambiguous and it's class label is set to $\mathbb{P}_\mathrm{label}(\text{Pneumonia opacity}) = 0.33$,  Otherwise, it is considered easy and $\mathbb{P}_\mathrm{label}(\text{Pneumonia opacity}) = 0$.
\end{itemize}

\subsection*{Learning noise-robust image representations.}
Noise-robust learning (NRL) is essential for handling noise and inconsistencies in labels which can severely impact the predictive performance of models \cite{arpit2017closer, zhang2016understanding}. Although, this type of machine learning is most commonly employed to improve model's predictive performance, in our setting, we use NRL methods to obtain an unbiased sample scoring function in active label cleaning. Interested readers should refer to the extended literature review \cite{song2020review_paper} for further information on this topic.

The sample selection algorithms presented in the next section heavily rely on models trained with NRL methods; in particular, we are utilising methods from two broad categories: model regularisation \cite{weight_decay, pmlr-lukasik20a, dropout_generalization} and sample exclusion approaches \cite{han2018co, huang2019o2u, yu19coteaching}. The former set of methods have been shown to be effective against memorising noisy labels \cite{arpit2017closer}, by favouring simpler decision boundaries. Sample exclusion approaches, on the other hand, aim to identify noisy samples during training by tracking loss values of individual samples \cite{han2018co, yu19coteaching} or gradient vector distributions \cite{liu2020early, parascandolo2020learning}. Alternative meta-model based approaches are not considered in this study as unbiased clean validation set may not always be available in real-world scenarios.

\subsection*{Sample selection algorithms.}
\label{sec:selection_algs}
Here we propose three selection algorithms for accurately estimating class posteriors, $p_{\boldsymbol{\theta}}(\hat{y} \vert \mathbf{x})$, to prioritise mislabelled data points in the scoring function $\Phi$ (see \cref{eqn:scoring_fn}).
In detail, we first train deep neural networks with noise robustness, then use them to identify the corrupted labels.
The methods differ in how the robustness is introduced and how the networks are used afterwards. These methods were proposed and studied to take into account different noise cleaning setups, including the number of labelled and unlabelled samples, and prior knowledge on the expected noise rate.

As a baseline approach, networks are trained with noisy labels and augmented images by minimising a negative log-likelihood loss term, which is referred to as vanilla. This approach is expected to be biased by the noisy labels and to perform sub-optimally in prioritising samples for relabelling.

\paragraph{Supervised co-teaching training.}
\label{sec:coteaching}
To cope with noisy labels, we first propose to use co-teaching \cite{han2018co} in a supervised learning setting, which has shown promising performance in classification accuracy and robustness.
In the co-teaching scheme, two identical networks $f_{\boldsymbol{\theta}_1}$ and $f_{\boldsymbol{\theta}_2}$ with different initialisation are trained simultaneously, but the batch of images at each training step for $f_{\boldsymbol{\theta}_1}$ is selected by $f_{\boldsymbol{\theta}_2}$ and vice versa.
The rationale is that images with high loss values during early training are ones that are difficult to learn, because their labels disagree with what the network has learned from the easy cases---indicating these labels might be corrupted.
Also, using two networks rather than one prevents a self-confirmation loop as in \cite{jiang2018mentornet}.
Co-teaching thus provides a method for robust training with noisy labels that can be employed to identify corrupted labels.
We ensemble the predictions of both trained networks as $\frac12 \sum_m f_{\boldsymbol{\theta}_m}(\mathbf{x}_i) = \hat{\boldsymbol{\pi}}_i$, where $\hat{\boldsymbol{\pi}}_i \in [0,1]^C$ and $\sum_{c=1}^C \hat{\pi}_i^c = 1$, and use the predicted distribution, $\hat{\boldsymbol{\pi}}_i$, in the scoring function $\Phi$ to rank the samples.

\paragraph{Self-supervision for robust learning.}
\label{sec:ssl_linear}
Noise-robust learning can still be influenced by label noise and exclude difficult samples, as they yield large loss values. Self-supervised learning (SSL) is proposed as a second approach to address this and further improve the sample ranking performance whenever an auxiliary task can be defined to pre-train models. As the approach does not require target labels and only extracts information from the images, it is an ideal way to learn domain-specific image encoders that are unbiased by the label noise. It can also be used in conjunction with the co-teaching algorithm in an end-to-end manner.

To this end we use BYOL \cite{grill2020bootstrap} to learn an embedding for the images, i.e., a non-linear transformation $\mathbf{x}_i \mapsto g_{\boldsymbol{\theta}}(\mathbf{x}_i) = \mathbf{h}_i$, where $\mathbf{h}_i \in \mathds{R}^L$ and $L$ is the size of the embedding space. As SSL is agnostic to the labels, the embedding will be formed based only on the image content, such that similar images will be close in the embedding space and dissimilar images will be further apart. The learnt encoder can later be used to build noise-robust models. One way to achieve this is by fixing the weights of $g_{\boldsymbol{\theta}}$ and train a linear classifier on top of the learned embedding as $\mathbf{h}_i \mapsto f_\varphi(\mathbf{h}_i) = \hat{\boldsymbol{\pi}}_i$, which is referred to as {``SSL-Linear"}. To obtain better calibrated posteriors, large logits are penalised at training time with function $\gamma(x) = \alpha \cdot \tanh(x/\alpha)$ \cite{grill2020bootstrap} and optimised with smoothed labels \cite{NEURIPS_labelsmoothing}. The linear classifier can only introduce a linear separation of the images according to their given labels. Assuming images with similar contents (i.e.\ nearby in embedding space) should have similar labels, the simple linear decision boundary introduces robustness to mislabels.

\section*{Data availability}
The CIFAR10H\cite{peterson2019human} dataset is available at \url{https://github.com/jcpeterson/cifar-10h}.
The raw data for NoisyCXR\ can be downloaded from the RSNA Pneumonia Detection Challenge\cite{KaggleRSNA}, the detailed annotations per annotator for this challenge can be downloaded from \href{https://storage.googleapis.com/kaggle-forum-message-attachments/844865/15542/RSNA_pneumonia_all_probs.csv.zip}{here} and the original labels from the NIH ChestXray dataset\cite{NIHDataset} can be found at \href{https://s3.amazonaws.com/east1.public.rsna.org/AI/2018/pneumonia-challenge-dataset-mappings_2018.json}{the link}. The labels used in our experiments are processed in the linked code repository \url{https://github.com/microsoft/InnerEye-DeepLearning/tree/main/InnerEye-DataQuality}. For further information on the acquisition and anonymisation of chest X-ray scans, and patients' informed consent, please see the online information \cite{nihcxr_public_release}. The authors have not collected any data themselves as the study only used public, widely used anonymised datasets. The National Institutes of Health (NIH) Clinical Center was responsible for the release of the NIH dataset images (also used for the RSNA Pneumonia Detection Challenge). Senior Investigator: Ronald M. Summers, M.D., Ph.D., Senior Investigator of the Clinical Image Processing Service in the Imaging Biomarkers and Computer-Aided Diagnosis Laboratory of the NIH Clinical Center Radiology and Imaging Sciences Department. Additionally, the proposed study was reviewed by Microsoft Research Ethics Review Program’s IRB, and met all the required ethical considerations outlined by the U.S. HHS Office for Human Research Protections, prior to initiation of the research (Reference IRB 10054, RCT 4018, 4375,4374). This was classified as non-human subjects research.

\section*{Code availability}
All the code for our label cleaning benchmarks, robust learning models, and experiments has been released in an open-source repository \url{https://github.com/microsoft/InnerEye-DeepLearning/tree/main/InnerEye-DataQuality}.  The repository contains all the configuration files and library requirements for the reproducibility of experiments presented in this study.

\bibliographystyle{naturemag}
\bibliography{references}

\begin{thebibliography}{10}
\expandafter\ifx\csname url\endcsname\relax
  \def\url#1{\texttt{#1}}\fi
\expandafter\ifx\csname urlprefix\endcsname\relax\def\urlprefix{URL }\fi
\providecommand{\bibinfo}[2]{#2}
\providecommand{\eprint}[2][]{\url{#2}}

\bibitem{northcutt_nips2020_workshop}
\bibinfo{author}{Northcutt, C.~G.}, \bibinfo{author}{Athalye, A.} \&
  \bibinfo{author}{Lin, J.}
\newblock \bibinfo{title}{Pervasive label errors in {ML} benchmark test sets,
  consequences, and benefits}.
\newblock In \emph{\bibinfo{booktitle}{NeurIPS 2020 Workshop on Security and
  Data Curation Workshop}} (\bibinfo{year}{2020}).

\bibitem{majkowska2020chest}
\bibinfo{author}{Majkowska, A.} \emph{et~al.}
\newblock \bibinfo{title}{Chest radiograph interpretation with deep learning
  models: assessment with radiologist-adjudicated reference standards and
  population-adjusted evaluation}.
\newblock \emph{\bibinfo{journal}{Radiology}} \textbf{\bibinfo{volume}{294}},
  \bibinfo{pages}{421--431} (\bibinfo{year}{2020}).

\bibitem{NIHDataset}
\bibinfo{author}{{Wang}, X.} \emph{et~al.}
\newblock \bibinfo{title}{{ChestX-Ray8}: Hospital-scale chest {X}-ray database
  and benchmarks on weakly-supervised classification and localization of common
  thorax diseases}.
\newblock In \emph{\bibinfo{booktitle}{2017 IEEE Conference on Computer Vision
  and Pattern Recognition (CVPR)}}, \bibinfo{pages}{3462--3471}
  (\bibinfo{year}{2017}).

\bibitem{beyer2020imagenet}
\bibinfo{author}{Beyer, L.}, \bibinfo{author}{Hénaff, O.~J.},
  \bibinfo{author}{Kolesnikov, A.}, \bibinfo{author}{Zhai, X.} \&
  \bibinfo{author}{van~den Oord, A.}
\newblock \bibinfo{title}{Are we done with {ImageNet}?}
\newblock \emph{\bibinfo{journal}{arXiv preprint arXiv:2006.07159}}
  (\bibinfo{year}{2020}).

\bibitem{peterson2019human}
\bibinfo{author}{Peterson, J.~C.}, \bibinfo{author}{Battleday, R.~M.},
  \bibinfo{author}{Griffiths, T.~L.} \& \bibinfo{author}{Russakovsky, O.}
\newblock \bibinfo{title}{Human uncertainty makes classification more robust}.
\newblock In \emph{\bibinfo{booktitle}{Proceedings of the IEEE International
  Conference on Computer Vision}}, \bibinfo{pages}{9617--9626}
  (\bibinfo{year}{2019}).

\bibitem{arpit2017closer}
\bibinfo{author}{Arpit, D.} \emph{et~al.}
\newblock \bibinfo{title}{A closer look at memorization in deep networks}.
\newblock In \emph{\bibinfo{booktitle}{Proceedings of the 34th International
  Conference on Machine Learning}}, vol.~\bibinfo{volume}{70} of
  \emph{\bibinfo{series}{PMLR}}, \bibinfo{pages}{233--242}
  (\bibinfo{publisher}{PMLR}, \bibinfo{year}{2017}).
\newblock \eprint{arXiv:1706.05394}.

\bibitem{zhang2016understanding}
\bibinfo{author}{Zhang, C.}, \bibinfo{author}{Bengio, S.},
  \bibinfo{author}{Hardt, M.}, \bibinfo{author}{Recht, B.} \&
  \bibinfo{author}{Vinyals, O.}
\newblock \bibinfo{title}{Understanding deep learning requires rethinking
  generalization}.
\newblock In \emph{\bibinfo{booktitle}{International Conference on Learning
  Representations}} (\bibinfo{year}{2017}).
\newblock \urlprefix\url{https://openreview.net/forum?id=Sy8gdB9xx}.

\bibitem{OAKDENRAYNER2020106}
\bibinfo{author}{Oakden-Rayner, L.}
\newblock \bibinfo{title}{Exploring large-scale public medical image datasets}.
\newblock \emph{\bibinfo{journal}{Academic Radiology}}
  \textbf{\bibinfo{volume}{27}}, \bibinfo{pages}{106 -- 112}
  (\bibinfo{year}{2020}).
\newblock
  \urlprefix\url{http://www.sciencedirect.com/science/article/pii/S107663321930488X}.
\newblock \bibinfo{note}{Special Issue: Artificial Intelligence}.

\bibitem{KaggleRSNA}
\bibinfo{title}{{RSNA} pneumonia detection challenge}.
\newblock
  \bibinfo{howpublished}{\url{https://www.kaggle.com/c/rsna-pneumonia-detection-challenge}}
  (\bibinfo{year}{2018}).

\bibitem{krause2018grader}
\bibinfo{author}{Krause, J.} \emph{et~al.}
\newblock \bibinfo{title}{Grader variability and the importance of reference
  standards for evaluating machine learning models for diabetic retinopathy}.
\newblock \emph{\bibinfo{journal}{Ophthalmology}}
  \textbf{\bibinfo{volume}{125}}, \bibinfo{pages}{1264--1272}
  (\bibinfo{year}{2018}).

\bibitem{gulshan2016development}
\bibinfo{author}{Gulshan, V.} \emph{et~al.}
\newblock \bibinfo{title}{Development and validation of a deep learning
  algorithm for detection of diabetic retinopathy in retinal fundus
  photographs}.
\newblock \emph{\bibinfo{journal}{JAMA}} \textbf{\bibinfo{volume}{316}},
  \bibinfo{pages}{2402--2410} (\bibinfo{year}{2016}).

\bibitem{battleday2020capturing}
\bibinfo{author}{Battleday, R.~M.}, \bibinfo{author}{Peterson, J.~C.} \&
  \bibinfo{author}{Griffiths, T.~L.}
\newblock \bibinfo{title}{Capturing human categorization of natural images by
  combining deep networks and cognitive models}.
\newblock \emph{\bibinfo{journal}{Nature Communications}}
  \textbf{\bibinfo{volume}{11}}, \bibinfo{pages}{1--14} (\bibinfo{year}{2020}).

\bibitem{li_dividemix}
\bibinfo{author}{Junnan~Li, S. C. H.~H., Richard~Socher}.
\newblock \bibinfo{title}{{DivideMix}: Learning with noisy labels as
  semi-supervised learning}.
\newblock In \emph{\bibinfo{booktitle}{International Conference on Learning
  Representations}} (\bibinfo{year}{2020}).

\bibitem{yu19coteaching}
\bibinfo{author}{Yu, X.} \emph{et~al.}
\newblock \bibinfo{title}{How does disagreement help generalization against
  label corruption?}
\newblock In \emph{\bibinfo{booktitle}{International Conference on Machine
  Learning}} (\bibinfo{year}{2019}).

\bibitem{han2018co}
\bibinfo{author}{Han, B.} \emph{et~al.}
\newblock \bibinfo{title}{Co-teaching: Robust training of deep neural networks
  with extremely noisy labels}.
\newblock In \emph{\bibinfo{booktitle}{Advances in Neural Information
  Processing Systems}}, \bibinfo{pages}{8527--8537} (\bibinfo{year}{2018}).

\bibitem{grill2020bootstrap}
\bibinfo{author}{Grill, J.-B.} \emph{et~al.}
\newblock \bibinfo{title}{Bootstrap your own latent - a new approach to
  self-supervised learning}.
\newblock In \emph{\bibinfo{booktitle}{Advances in Neural Information
  Processing Systems}}, vol.~\bibinfo{volume}{33},
  \bibinfo{pages}{21271--21284} (\bibinfo{year}{2020}).

\bibitem{reactive_learning}
\bibinfo{author}{Lin, C.~H.}, \bibinfo{author}{Mausam} \&
  \bibinfo{author}{Weld, D.~S.}
\newblock \bibinfo{title}{Re-active learning: Active learning with relabeling}.
\newblock In \emph{\bibinfo{booktitle}{Proceedings of the Thirtieth AAAI
  Conference on Artificial Intelligence}}, \bibinfo{pages}{1845–1852}
  (\bibinfo{publisher}{AAAI Press}, \bibinfo{year}{2016}).

\bibitem{ipeirotis2014repeated}
\bibinfo{author}{Ipeirotis, P.~G.}, \bibinfo{author}{Provost, F.},
  \bibinfo{author}{Sheng, V.~S.} \& \bibinfo{author}{Wang, J.}
\newblock \bibinfo{title}{Repeated labeling using multiple noisy labelers}.
\newblock \emph{\bibinfo{journal}{Data Mining and Knowledge Discovery}}
  \textbf{\bibinfo{volume}{28}}, \bibinfo{pages}{402--441}
  (\bibinfo{year}{2014}).

\bibitem{ekambaram2016active}
\bibinfo{author}{Ekambaram, R.} \emph{et~al.}
\newblock \bibinfo{title}{Active cleaning of label noise}.
\newblock \emph{\bibinfo{journal}{Pattern Recognition}}
  \textbf{\bibinfo{volume}{51}}, \bibinfo{pages}{463--480}
  (\bibinfo{year}{2016}).

\bibitem{northcutt2019confident}
\bibinfo{author}{Northcutt, C.~G.}, \bibinfo{author}{Jiang, L.} \&
  \bibinfo{author}{Chuang, I.~L.}
\newblock \bibinfo{title}{Confident learning: Estimating uncertainty in dataset
  labels}.
\newblock \emph{\bibinfo{journal}{arXiv preprint arXiv:1911.00068}}
  (\bibinfo{year}{2019}).

\bibitem{huang2019o2u}
\bibinfo{author}{Huang, J.}, \bibinfo{author}{Qu, L.}, \bibinfo{author}{Jia,
  R.} \& \bibinfo{author}{Zhao, B.}
\newblock \bibinfo{title}{{O2U-Net}: A simple noisy label detection approach
  for deep neural networks}.
\newblock In \emph{\bibinfo{booktitle}{Proceedings of the IEEE/CVF
  International Conference on Computer Vision}}, \bibinfo{pages}{3326--3334}
  (\bibinfo{year}{2019}).

\bibitem{zhang_generalized_cross_entropy}
\bibinfo{author}{Zhang, Z.} \& \bibinfo{author}{Sabuncu, M.}
\newblock \bibinfo{title}{Generalized cross entropy loss for training deep
  neural networks with noisy labels}.
\newblock In \emph{\bibinfo{booktitle}{Advances in Neural Information
  Processing Systems 31}}, \bibinfo{pages}{8778--8788} (\bibinfo{year}{2018}).

\bibitem{houlsby2011bald}
\bibinfo{author}{Houlsby, N.}, \bibinfo{author}{Huszár, F.},
  \bibinfo{author}{Ghahramani, Z.} \& \bibinfo{author}{Lengyel, M.}
\newblock \bibinfo{title}{{B}ayesian active learning for classification and
  preference learning} (\bibinfo{year}{2011}).
\newblock \urlprefix\url{https://arxiv.org/abs/1112.5745}.

\bibitem{kirsch2019batchbald}
\bibinfo{author}{Kirsch, A.}, \bibinfo{author}{Van~Amersfoort, J.} \&
  \bibinfo{author}{Gal, Y.}
\newblock \bibinfo{title}{Batchbald: Efficient and diverse batch acquisition
  for deep bayesian active learning}.
\newblock \emph{\bibinfo{journal}{Advances in neural information processing
  systems}} \textbf{\bibinfo{volume}{32}} (\bibinfo{year}{2019}).

\bibitem{grandvalet2004semi}
\bibinfo{author}{Grandvalet, Y.} \& \bibinfo{author}{Bengio, Y.}
\newblock \bibinfo{title}{Semi-supervised learning by entropy minimization}.
\newblock In \emph{\bibinfo{booktitle}{Proceedings of the 17th International
  Conference on Neural Information Processing Systems}},
  \bibinfo{pages}{529--536} (\bibinfo{year}{2004}).

\bibitem{pereyra2017regularizing}
\bibinfo{author}{Pereyra, G.}, \bibinfo{author}{Tucker, G.},
  \bibinfo{author}{Chorowski, J.}, \bibinfo{author}{Kaiser, {\L}.} \&
  \bibinfo{author}{Hinton, G.}
\newblock \bibinfo{title}{Regularizing neural networks by penalizing confident
  output distributions}.
\newblock \emph{\bibinfo{journal}{arXiv preprint arXiv:1701.06548}}
  (\bibinfo{year}{2017}).

\bibitem{krizhevsky2009learning}
\bibinfo{author}{Krizhevsky, A.}
\newblock \bibinfo{title}{Learning multiple layers of features from tiny
  images}.
\newblock \bibinfo{type}{Tech. Rep.}, \bibinfo{institution}{University of
  Toronto} (\bibinfo{year}{2009}).

\bibitem{xia2020part}
\bibinfo{author}{Xia, X.} \emph{et~al.}
\newblock \bibinfo{title}{Part-dependent label noise: Towards
  instance-dependent label noise}.
\newblock In \emph{\bibinfo{booktitle}{Advances in Neural Information
  Processing Systems}}, vol.~\bibinfo{volume}{33} (\bibinfo{year}{2020}).

\bibitem{gal2017deep}
\bibinfo{author}{Gal, Y.}, \bibinfo{author}{Islam, R.} \&
  \bibinfo{author}{Ghahramani, Z.}
\newblock \bibinfo{title}{Deep {B}ayesian active learning with image data}.
\newblock In \emph{\bibinfo{booktitle}{International Conference on Machine
  Learning}}, \bibinfo{pages}{1183--1192} (\bibinfo{publisher}{PMLR},
  \bibinfo{year}{2017}).

\bibitem{liu2020early}
\bibinfo{author}{Liu, S.}, \bibinfo{author}{Niles-Weed, J.},
  \bibinfo{author}{Razavian, N.} \& \bibinfo{author}{Fernandez-Granda, C.}
\newblock \bibinfo{title}{Early-learning regularization prevents memorization
  of noisy labels}.
\newblock \emph{\bibinfo{journal}{Advances in Neural Information Processing
  Systems}} \textbf{\bibinfo{volume}{33}} (\bibinfo{year}{2020}).

\bibitem{chen2020robustness}
\bibinfo{author}{Chen, P.}, \bibinfo{author}{Ye, J.}, \bibinfo{author}{Chen,
  G.}, \bibinfo{author}{Zhao, J.} \& \bibinfo{author}{Heng, P.-A.}
\newblock \bibinfo{title}{Robustness of accuracy metric and its inspirations in
  learning with noisy labels}.
\newblock In \emph{\bibinfo{booktitle}{Proceedings of the AAAI Conference on
  Artificial Intelligence}}, vol.~\bibinfo{volume}{35},
  \bibinfo{pages}{11451--11461} (\bibinfo{year}{2021}).

\bibitem{patrini2017making}
\bibinfo{author}{Patrini, G.}, \bibinfo{author}{Rozza, A.},
  \bibinfo{author}{Krishna~Menon, A.}, \bibinfo{author}{Nock, R.} \&
  \bibinfo{author}{Qu, L.}
\newblock \bibinfo{title}{Making deep neural networks robust to label noise: A
  loss correction approach}.
\newblock In \emph{\bibinfo{booktitle}{Proceedings of the IEEE Conference on
  Computer Vision and Pattern Recognition (CVPR)}}, \bibinfo{pages}{1944--1952}
  (\bibinfo{year}{2017}).

\bibitem{toneva2018empirical}
\bibinfo{author}{Toneva, M.} \emph{et~al.}
\newblock \bibinfo{title}{An empirical study of example forgetting during deep
  neural network learning}.
\newblock In \emph{\bibinfo{booktitle}{International Conference on Learning
  Representations}} (\bibinfo{year}{2018}).

\bibitem{sener2018active}
\bibinfo{author}{Sener, O.} \& \bibinfo{author}{Savarese, S.}
\newblock \bibinfo{title}{Active learning for convolutional neural networks: A
  core-set approach}.
\newblock In \emph{\bibinfo{booktitle}{International Conference on Learning
  Representations}} (\bibinfo{year}{2018}).

\bibitem{nguyen2020self}
\bibinfo{author}{Nguyen, D.~T.} \emph{et~al.}
\newblock \bibinfo{title}{{SELF}: Learning to filter noisy labels with
  self-ensembling}.
\newblock In \emph{\bibinfo{booktitle}{International Conference on Learning
  Representations}} (\bibinfo{year}{2020}).

\bibitem{lam2003evaluating}
\bibinfo{author}{Lam, C.~P.} \& \bibinfo{author}{Stork, D.~G.}
\newblock \bibinfo{title}{Evaluating classifiers by means of test data with
  noisy labels}.
\newblock In \emph{\bibinfo{booktitle}{IJCAI}}, vol.~\bibinfo{volume}{3},
  \bibinfo{pages}{513--518} (\bibinfo{year}{2003}).

\bibitem{he2016deep}
\bibinfo{author}{He, K.}, \bibinfo{author}{Zhang, X.}, \bibinfo{author}{Ren,
  S.} \& \bibinfo{author}{Sun, J.}
\newblock \bibinfo{title}{Deep residual learning for image recognition}.
\newblock In \emph{\bibinfo{booktitle}{Proceedings of the IEEE Conference on
  Computer Vision and Pattern Recognition (CVPR)}}, \bibinfo{pages}{770--778}
  (\bibinfo{year}{2016}).

\bibitem{cheng2020learning}
\bibinfo{author}{Cheng, J.}, \bibinfo{author}{Liu, T.},
  \bibinfo{author}{Ramamohanarao, K.} \& \bibinfo{author}{Tao, D.}
\newblock \bibinfo{title}{Learning with bounded instance and label-dependent
  label noise}.
\newblock In \emph{\bibinfo{booktitle}{International Conference on Machine
  Learning}}, \bibinfo{pages}{1789--1799} (\bibinfo{organization}{PMLR},
  \bibinfo{year}{2020}).

\bibitem{khetan2017learning}
\bibinfo{author}{Khetan, A.}, \bibinfo{author}{Lipton, Z.~C.} \&
  \bibinfo{author}{Anandkumar, A.}
\newblock \bibinfo{title}{Learning from noisy singly-labeled data}.
\newblock In \emph{\bibinfo{booktitle}{International Conference on Learning
  Representations}} (\bibinfo{year}{2018}).
\newblock \eprint{arXiv:1712.04577}.

\bibitem{rodrigues2013learning}
\bibinfo{author}{Rodrigues, F.}, \bibinfo{author}{Pereira, F.} \&
  \bibinfo{author}{Ribeiro, B.}
\newblock \bibinfo{title}{Learning from multiple annotators: distinguishing
  good from random labelers}.
\newblock \emph{\bibinfo{journal}{Pattern Recognition Letters}}
  \textbf{\bibinfo{volume}{34}}, \bibinfo{pages}{1428--1436}
  (\bibinfo{year}{2013}).

\bibitem{tanno2019learning}
\bibinfo{author}{Tanno, R.}, \bibinfo{author}{Saeedi, A.},
  \bibinfo{author}{Sankaranarayanan, S.}, \bibinfo{author}{Alexander, D.~C.} \&
  \bibinfo{author}{Silberman, N.}
\newblock \bibinfo{title}{Learning from noisy labels by regularized estimation
  of annotator confusion}.
\newblock In \emph{\bibinfo{booktitle}{Proceedings of the IEEE/CVF Conference
  on Computer Vision and Pattern Recognition (CVPR)}},
  \bibinfo{pages}{11244--11253} (\bibinfo{year}{2019}).

\bibitem{warfield2004simultaneous}
\bibinfo{author}{Warfield, S.~K.}, \bibinfo{author}{Zou, K.~H.} \&
  \bibinfo{author}{Wells, W.~M.}
\newblock \bibinfo{title}{Simultaneous truth and performance level estimation
  ({STAPLE}): An algorithm for the validation of image segmentation}.
\newblock \emph{\bibinfo{journal}{IEEE Transactions on Medical Imaging}}
  \textbf{\bibinfo{volume}{23}}, \bibinfo{pages}{903--921}
  (\bibinfo{year}{2004}).

\bibitem{frenay2014classification}
\bibinfo{author}{Fr{\'e}nay, B.} \& \bibinfo{author}{Verleysen, M.}
\newblock \bibinfo{title}{Classification in the presence of label noise: a
  survey}.
\newblock \emph{\bibinfo{journal}{IEEE Transactions on Neural Networks and
  Learning Systems}} \textbf{\bibinfo{volume}{25}}, \bibinfo{pages}{845--869}
  (\bibinfo{year}{2014}).

\bibitem{Guo2017}
\bibinfo{author}{Guo, C.}, \bibinfo{author}{Pleiss, G.}, \bibinfo{author}{Sun,
  Y.} \& \bibinfo{author}{Weinberger, K.~Q.}
\newblock \bibinfo{title}{On calibration of modern neural networks}.
\newblock In \emph{\bibinfo{booktitle}{International Conference on Machine
  Learning}}, \bibinfo{pages}{1321–1330} (\bibinfo{year}{2017}).

\bibitem{irvin2019chexpert}
\bibinfo{author}{Irvin, J.} \emph{et~al.}
\newblock \bibinfo{title}{{CheXpert}: A large chest radiograph dataset with
  uncertainty labels and expert comparison}.
\newblock In \emph{\bibinfo{booktitle}{Proceedings of the AAAI Conference on
  Artificial Intelligence}}, vol.~\bibinfo{volume}{33},
  \bibinfo{pages}{590--597} (\bibinfo{year}{2019}).

\bibitem{oakdenrayner2017thoughts}
\bibinfo{author}{Oakden-Rayner, L.}
\newblock \bibinfo{title}{Quick thoughts on {ChestXray14}, performance claims,
  and clinical tasks} (\bibinfo{year}{2017}).
\newblock
  \urlprefix\url{https://lukeoakdenrayner.wordpress.com/2017/11/18/quick-thoughts-on-chestxray14-performance-claims-and-clinical-tasks/}.

\bibitem{kagglecxrboxes}
\bibinfo{author}{Stein, A.}
\newblock \bibinfo{title}{Pneumonia boxes likelihood data -- {RSNA} pneumonia
  detection challenge} (\bibinfo{year}{2020}).
\newblock
  \urlprefix\url{https://www.kaggle.com/c/rsna-pneumonia-detection-challenge/discussion/150643}.

\bibitem{song2020review_paper}
\bibinfo{author}{Song, H.}, \bibinfo{author}{Kim, M.}, \bibinfo{author}{Park,
  D.} \& \bibinfo{author}{Lee, J.-G.}
\newblock \bibinfo{title}{Learning from noisy labels with deep neural networks:
  A survey}.
\newblock \emph{\bibinfo{journal}{arXiv preprint arXiv:2007.08199}}
  (\bibinfo{year}{2020}).

\bibitem{weight_decay}
\bibinfo{author}{Krogh, A.} \& \bibinfo{author}{Hertz, J.~A.}
\newblock \bibinfo{title}{A simple weight decay can improve generalization}.
\newblock In \emph{\bibinfo{booktitle}{Advances in Neural Information
  Processing Systems 4}}, \bibinfo{pages}{950--957}
  (\bibinfo{publisher}{Morgan-Kaufmann}, \bibinfo{year}{1992}).

\bibitem{pmlr-lukasik20a}
\bibinfo{author}{Lukasik, M.}, \bibinfo{author}{Bhojanapalli, S.},
  \bibinfo{author}{Menon, A.} \& \bibinfo{author}{Kumar, S.}
\newblock \bibinfo{title}{Does label smoothing mitigate label noise?}
\newblock In \emph{\bibinfo{booktitle}{Proceedings of the 37th International
  Conference on Machine Learning}}, \bibinfo{pages}{6448--6458}
  (\bibinfo{year}{2020}).

\bibitem{dropout_generalization}
\bibinfo{author}{Srivastava, N.}, \bibinfo{author}{Hinton, G.},
  \bibinfo{author}{Krizhevsky, A.}, \bibinfo{author}{Sutskever, I.} \&
  \bibinfo{author}{Salakhutdinov, R.}
\newblock \bibinfo{title}{Dropout: A simple way to prevent neural networks from
  overfitting}.
\newblock \emph{\bibinfo{journal}{Journal of Machine Learning Research}}
  \textbf{\bibinfo{volume}{15}}, \bibinfo{pages}{1929--1958}
  (\bibinfo{year}{2014}).

\bibitem{parascandolo2020learning}
\bibinfo{author}{Parascandolo, G.}, \bibinfo{author}{Neitz, A.},
  \bibinfo{author}{Orvieto, A.}, \bibinfo{author}{Gresele, L.} \&
  \bibinfo{author}{Sch{\"o}lkopf, B.}
\newblock \bibinfo{title}{Learning explanations that are hard to vary}.
\newblock \emph{\bibinfo{journal}{arXiv preprint arXiv:2009.00329}}
  (\bibinfo{year}{2020}).

\bibitem{jiang2018mentornet}
\bibinfo{author}{Jiang, L.}, \bibinfo{author}{Zhou, Z.},
  \bibinfo{author}{Leung, T.}, \bibinfo{author}{Li, L.-J.} \&
  \bibinfo{author}{Fei-Fei, L.}
\newblock \bibinfo{title}{{MentorNet}: Learning data-driven curriculum for very
  deep neural networks on corrupted labels}.
\newblock In \emph{\bibinfo{booktitle}{International Conference on Machine
  Learning}}, \bibinfo{pages}{2309--2318} (\bibinfo{year}{2018}).

\bibitem{NEURIPS_labelsmoothing}
\bibinfo{author}{M\"{u}ller, R.}, \bibinfo{author}{Kornblith, S.} \&
  \bibinfo{author}{Hinton, G.~E.}
\newblock \bibinfo{title}{When does label smoothing help?}
\newblock In \emph{\bibinfo{booktitle}{Advances in Neural Information
  Processing Systems}}, vol.~\bibinfo{volume}{32} (\bibinfo{year}{2019}).

\bibitem{nihcxr_public_release}
\bibinfo{title}{{NIH} clinical center provides one of the largest publicly
  available chest x-ray datasets to scientific community}
  (\bibinfo{year}{2017}).
\newblock
  \urlprefix\url{https://www.nih.gov/news-events/news-releases/nih-clinical-center-provides-one-largest-publicly-available-chest-x-ray-datasets-scientific-community}.
\newblock \bibinfo{note}{Accessed: 2021-11-19}.

\end{thebibliography}


\begin{thebibliography}{10}
\expandafter\ifx\csname url\endcsname\relax
  \def\url#1{\texttt{#1}}\fi
\expandafter\ifx\csname urlprefix\endcsname\relax\def\urlprefix{URL }\fi
\providecommand{\bibinfo}[2]{#2}
\providecommand{\eprint}[2][]{\url{#2}}

\bibitem{frenay2014classification}
\bibinfo{author}{Fr{\'e}nay, B.} \& \bibinfo{author}{Verleysen, M.}
\newblock \bibinfo{title}{Classification in the presence of label noise: a
  survey}.
\newblock \emph{\bibinfo{journal}{IEEE Transactions on Neural Networks and
  Learning Systems}} \textbf{\bibinfo{volume}{25}}, \bibinfo{pages}{845--869}
  (\bibinfo{year}{2014}).

\bibitem{gal2017deep}
\bibinfo{author}{Gal, Y.}, \bibinfo{author}{Islam, R.} \&
  \bibinfo{author}{Ghahramani, Z.}
\newblock \bibinfo{title}{Deep {B}ayesian active learning with image data}.
\newblock In \emph{\bibinfo{booktitle}{International Conference on Machine
  Learning}}, \bibinfo{pages}{1183--1192} (\bibinfo{publisher}{PMLR},
  \bibinfo{year}{2017}).

\bibitem{houlsby2011bald}
\bibinfo{author}{Houlsby, N.}, \bibinfo{author}{Huszár, F.},
  \bibinfo{author}{Ghahramani, Z.} \& \bibinfo{author}{Lengyel, M.}
\newblock \bibinfo{title}{{B}ayesian active learning for classification and
  preference learning} (\bibinfo{year}{2011}).
\newblock \urlprefix\url{https://arxiv.org/abs/1112.5745}.

\bibitem{chen2020beyond}
\bibinfo{author}{Chen, P.}, \bibinfo{author}{Ye, J.}, \bibinfo{author}{Chen,
  G.}, \bibinfo{author}{Zhao, J.} \& \bibinfo{author}{Heng, P.-A.}
\newblock \bibinfo{title}{Beyond class-conditional assumption: A primary
  attempt to combat instance-dependent label noise}.
\newblock In \emph{\bibinfo{booktitle}{Proceedings of the AAAI Conference on
  Artificial Intelligence}}, vol.~\bibinfo{volume}{35}.

\bibitem{xia2020part}
\bibinfo{author}{Xia, X.} \emph{et~al.}
\newblock \bibinfo{title}{Part-dependent label noise: Towards
  instance-dependent label noise}.
\newblock In \emph{\bibinfo{booktitle}{Advances in Neural Information
  Processing Systems}}, vol.~\bibinfo{volume}{33} (\bibinfo{year}{2020}).

\bibitem{han2018co}
\bibinfo{author}{Han, B.} \emph{et~al.}
\newblock \bibinfo{title}{Co-teaching: Robust training of deep neural networks
  with extremely noisy labels}.
\newblock In \emph{\bibinfo{booktitle}{Advances in Neural Information
  Processing Systems}}, \bibinfo{pages}{8527--8537} (\bibinfo{year}{2018}).

\bibitem{grill2020bootstrap}
\bibinfo{author}{Grill, J.-B.} \emph{et~al.}
\newblock \bibinfo{title}{Bootstrap your own latent - a new approach to
  self-supervised learning}.
\newblock In \emph{\bibinfo{booktitle}{Advances in Neural Information
  Processing Systems}}, vol.~\bibinfo{volume}{33},
  \bibinfo{pages}{21271--21284} (\bibinfo{year}{2020}).

\bibitem{chen_simclr}
\bibinfo{author}{Chen, T.}, \bibinfo{author}{Kornblith, S.},
  \bibinfo{author}{Norouzi, M.} \& \bibinfo{author}{Hinton, G.}
\newblock \bibinfo{title}{A simple framework for contrastive learning of visual
  representations}.
\newblock In \emph{\bibinfo{booktitle}{Proceedings of the 37th International
  Conference on Machine Learning}}, \bibinfo{pages}{1597--1607}
  (\bibinfo{publisher}{PMLR}, \bibinfo{year}{2020}).

\bibitem{kingma2014adam}
\bibinfo{author}{Kingma, D.~P.} \& \bibinfo{author}{Ba, J.}
\newblock \bibinfo{title}{Adam: A method for stochastic optimization}.
\newblock In \emph{\bibinfo{booktitle}{International Conference on Learning
  Representations}} (\bibinfo{year}{2015}).
\newblock \eprint{arXiv:1412.6980}.

\bibitem{NIHDataset}
\bibinfo{author}{{Wang}, X.} \emph{et~al.}
\newblock \bibinfo{title}{{ChestX-Ray8}: Hospital-scale chest {X}-ray database
  and benchmarks on weakly-supervised classification and localization of common
  thorax diseases}.
\newblock In \emph{\bibinfo{booktitle}{2017 IEEE Conference on Computer Vision
  and Pattern Recognition (CVPR)}}, \bibinfo{pages}{3462--3471}
  (\bibinfo{year}{2017}).

\bibitem{simard2003best}
\bibinfo{author}{Simard, P.~Y.}, \bibinfo{author}{Steinkraus, D.} \&
  \bibinfo{author}{Platt, J.~C.}
\newblock \bibinfo{title}{Best practices for convolutional neural networks
  applied to visual document analysis}.
\newblock In \emph{\bibinfo{booktitle}{Proceedings of the Seventh International
  Conference on Document Analysis and Recognition (ICDAR 2003)}},
  \bibinfo{pages}{958--963} (\bibinfo{year}{2003}).

\end{thebibliography}

\section*{Acknowledgments}
This work was funded by Microsoft Research Ltd (Cambridge, UK). The authors would also like to extend their thanks to Hannah Murfet for guidance offered as part the compliance review of the datasets used in this study.

\section*{Author Contributions}
O.O., D.C.C., R.T., M.B. conceptualised and designed the benchmarks and methods presented in this manuscript. D.C.C., O.O, M.B., A.S. drafted the manuscript. D.C.C., O.O., M.B., J.A.V., R.T., S.B., M.L., A.N., B.G. revised the manuscript and gave critical feedback for important intellectual content. M.B., O.O., K.C.T., M.M. analysed the data. M.B., O.O., A.S., K.C.T., M.M. performed the experiments and statistical analysis. J.A.V. and A.N. secured funding. O.O., J.A.V., D.C.C., R.T. and A.N. supervised the project.

\section*{Competing Interests}
The authors declare no competing interests.

\end{document}


\captionsetup[figure]{name={Supp.~Fig.}}
\captionsetup[table]{name={Supp.~Table}}
\crefname{figure}{supplementary Fig.}{supplementary Figs.}
\crefname{table}{supplementary Table}{supplementary Tables}

\twocolumn[{
\begin{@twocolumnfalse}
\maketitle
\vspace*{1cm}
\end{@twocolumnfalse}}]


\section{Supplementary Note 1: Additional experiments}

\subsection{Classification performance of noise-robust models.}
\label{sec:classification_accuracy}
We test the robustness of the proposed approaches to the corrupted labels in the training set and measure how well they generalise to an unseen set with clean labels. To this end, all networks are trained on a dataset $\Dtrain = \{(\*x_i, y_i)\}_{i=1}^N$ containing both mislabelled and correctly labelled data points. 
To assess the robustness to the noisy labels, we separate $\Dtrain$ into disjoint subsets of correctly labelled ($\Dcorrect$) and mislabelled cases ($\Dmislab$).
Similarly, the robustness is measured on a hold-out evaluation set ($\Dval$) with clean labels. 

We analyse the classification metrics for these three groups separately throughout training for the different proposed methods. The results are reported in \cref{tab:classification_accuracy_of_models} in terms of multi-class classification accuracy (CIFAR10H) and ROC-AUC values for binary classification (\NoisyCXR). In this analysis setup, robustness means that the network should learn to ignore the incorrect labels of the mislabelled cases and yield an accuracy close to chance for these. On the other hand, this should not come at a price of reduced performance for the correctly labelled cases and the network should yield a high accuracy for these. Hence the ideal case would be a large gap between the accuracy of these two groups ($\Dcorrect$, $\Dmislab$) throughout training and the gap is a good proxy measure of sample selectors' performance in the relabelling procedure.

\subsection{Simulation experiments with larger noise rates.}
\label{sec:simulation_larger_noise_rates}
\paragraph{CIFAR10H}
Label cleaning simulations are repeated for larger noise rates to understand the sensitivity of selection algorithms to noise rate. This way we also explore the limits at which data-driven label cleaning algorithms may become unreliable. For this purpose, we use the same set of CIFAR10H images ($|\Dataset|=10k$) but with a different initial set of noisy labels sampled from the label distribution using a larger temperature parameter ($\tau > 2$). In this way, we are able to control the initial noise rate up to $34\%$. For noise values beyond this rate, a second parameter is introduced as an additive bias term applied to the confusion matrix, which can serve as a uniform symmetric or class-dependent noise model. At the end, we construct three different sets of initial labels with the following average noise rates and noise models: (I) 30\% noise (CIFAR10H distribution), (II) 50\% noise (CIFAR10H + symmetric), and (III) 50\% (CIFAR10H + class-dependent).

\begin{table}[t!]
    \tablestyle
    \scriptsize
	\caption{The models used in the simulation experiments are also evaluated in terms of their classification performance on images with clean and noisy labels.}
	\label{tab:classification_accuracy_of_models}
	\setlength{\tabcolsep}{3pt}
	\begin{tabular}{@{\extracolsep{1pt}}llccccccc@{}}
	    \toprule
		Dataset & Model & Acc on $\Dval$ & Acc on $\Dcorrect$ & Acc on $\Dmislab$ \\
		\midrule
		{\scriptsize\multirow{3}{*}{\textit{CIFAR10H}}}
		& Vanilla         & 77.22 (0.29) & 92.38 (0.20) &  45.50 (2.78)   \\
		& Co-teaching     & 79.19 (0.22) & 88.38 (0.26) &  26.79 (1.64)   \\
		& SSL + Linear head & 80.56 (0.11) & 85.37 (0.20) &  12.62 (0.44) \\
		\midrule
		& & AUC on $\Dval$ & AUC on $\Dcorrect$ & AUC on $\Dmislab$ \\
		\midrule
		{\scriptsize\multirow{3}{*}{\textit{\NoisyCXR}}}&
		Vanilla     & 83.52 (0.85)    &   89.87 (0.42)  &  32.71 (1.86)  \\
		& Co-teaching  & 84.91 (0.31)    &  90.17 (0.26)  &  28.10 (0.29) \\
        & SSL + Co-teaching      & 87.45 (0.02)  &   91.84 (0.16)  &  25.92 (0.19)  \\
		\bottomrule
	\end{tabular}
	\scriptsize \\[1.5 mm]
	The networks are trained on a dataset containing partly mislabelled images: CIFAR10H: $|\Dtrain| = 10k,\, \eta = 15\%$; \NoisyCXR: $|\Dtrain| = 13.3k,\, \eta=12.7\%$, where $\eta$ is the noise rate. Their performance is evaluated on an evaluation set $\Dval$ with clean labels, and training set with clean $\Dcorrect$ and noisy labels  $\Dmislab$ ($\Dmislab \cap \Dcorrect = \emptyset$). The size of the evaluation sets is $|\Dval| = 50k$ for CIFAR10H and $|\Dval| = 13.3k$ for \NoisyCXR. The results are aggregated over three separate runs with different seeds (std.\ deviation in parentheses) and reported in terms of classification accuracy (Acc) and ROC-AUC (AUC).
\end{table}

\begin{figure*}[t!]
    \centering
    \figstyle
    \begin{subfigure}{0.455\linewidth}
        \includegraphics[width=\linewidth]{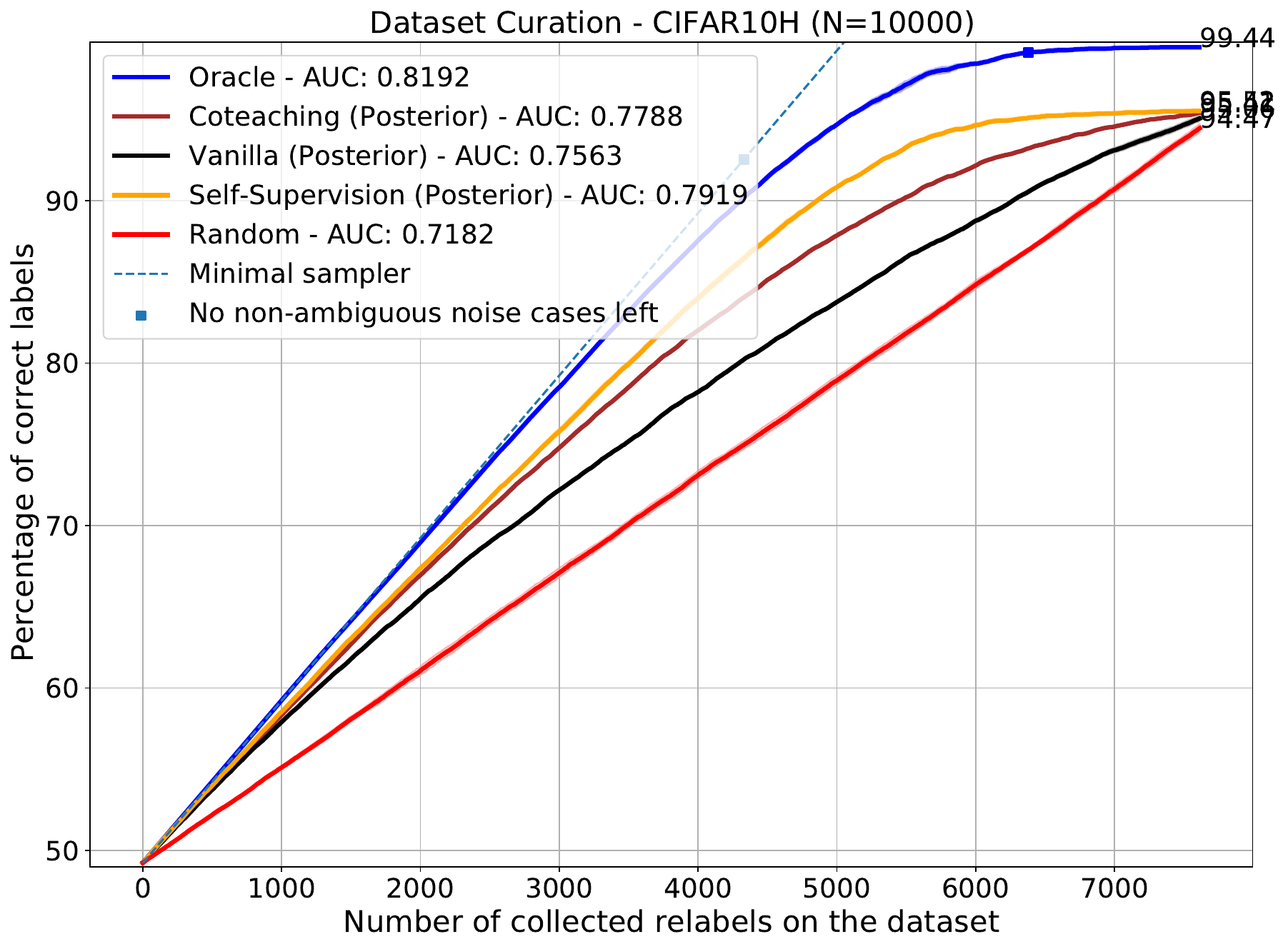}
        \caption{CIFAR10H -- Noise rate 50\% Symmetric Noise}
        \label{fig:CIFAR_50perc_sym}
    \end{subfigure}
    \begin{subfigure}{0.455\linewidth}
        \includegraphics[width=\linewidth]{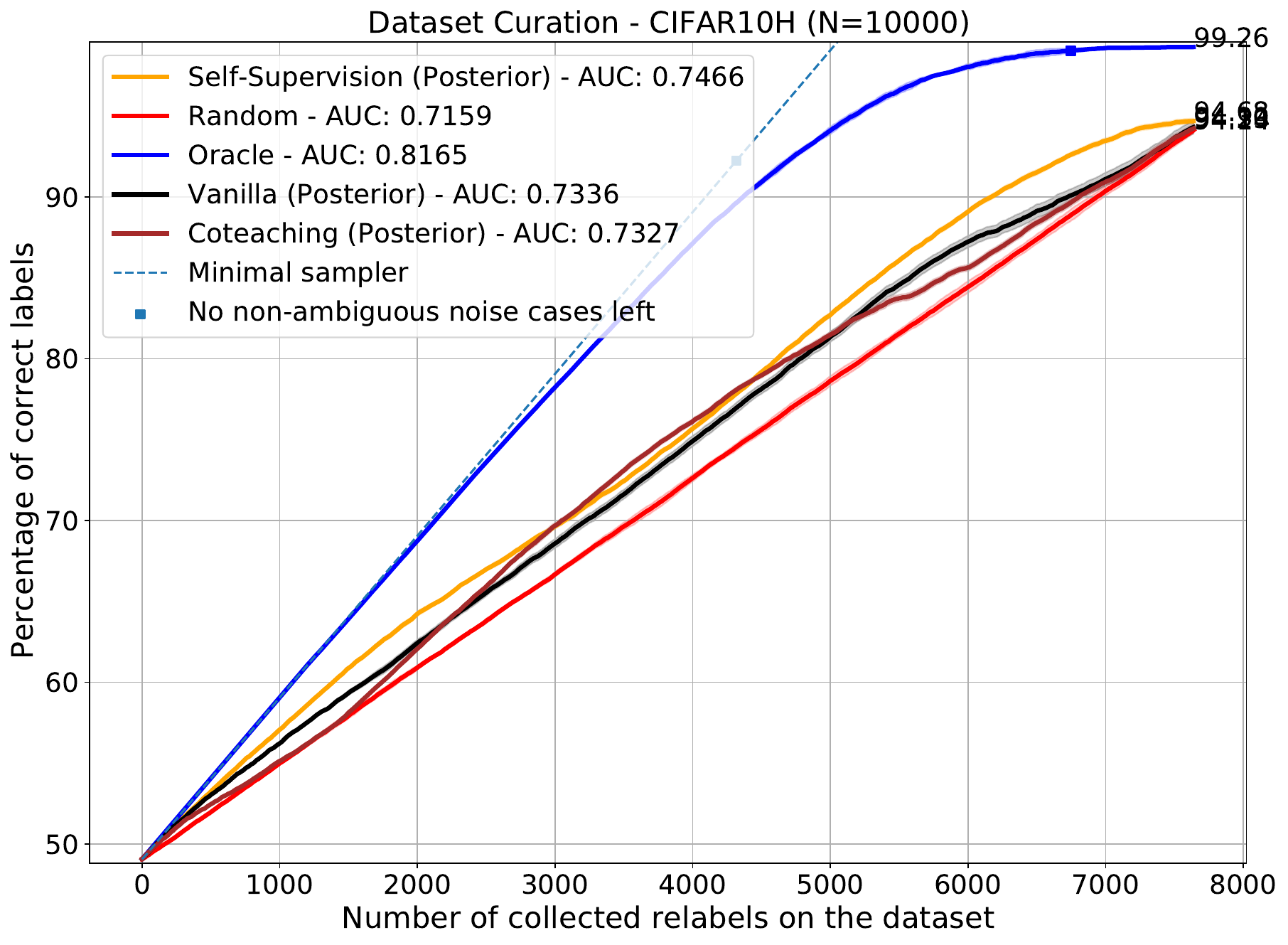}
        \caption{CIFAR10H -- Noise rate 50\% Asymmetric Noise}
        \label{fig:CIFAR_50perc_asym}
    \end{subfigure}
    \caption{Label cleaning performance for noise rate of 50\%, using symmetric noise and a more realistic, asymmetric noise model. Shaded areas represent +/- standard deviation over 5 random seeds for relabelling.}
\end{figure*}

For larger noise rates, the cost-effectiveness of sample selection algorithms are experimentally observed to be preserved as shown in \cref{fig:CIFAR_50perc_sym,fig:CIFAR_50perc_asym}. It is important to note that in 30\% noise case the breakdown of clear and difficult noisy cases are 23.24\% and 6.66\% respectively. On the other hand, when we switch to class-dependent noise model with more than 50\% noise rate, the sample ranking quality begins to degrade as diagonal dominance is no longer preserved for some certain classes (e.g.\ dog, cat, deer, horse, and automobile). We expect the algorithms to degrade towards random sampling as we introduce further bias and noise into the initial set of labels.

\paragraph{\NoisyCXR}
In this section, we investigate the impact of higher noise rates on label cleaning simulations for \NoisyCXR. To construct datasets with more noise, we followed the same procedure as described in Methods but instead of sampling 10\% of noise within the NIH ``Consolidation/Infiltration" category we sampled 25\% noise within this category. Including both the NIH noise and the sampled noise, the dataset contained in total 20\% noise. On this dataset, note that 43\% of positive initial (noisy) labels are incorrectly labelled, making it inherently challenging to learn an accurate classifier on this noisy training dataset. With this higher noise rate, we observe that the sample selection quality of \textit{co-teaching} degrade towards the \textit{vanilla} model. On the other hand, we still observe that combining co-teaching with self-supervised pre-training improves the quality of the label cleaning procedure, as the model is able to separate noisy from clean cases meaningfully earlier in training, hence improving co-teaching.

\subsection{Further analysis on the scoring function.}
\label{sec:scoring_fn_analysis}
\begin{figure*}[t]
    \centering
    \figstyle
    \begin{subfigure}{\linewidth}
        \centering
        \includegraphics[width=0.32\linewidth]{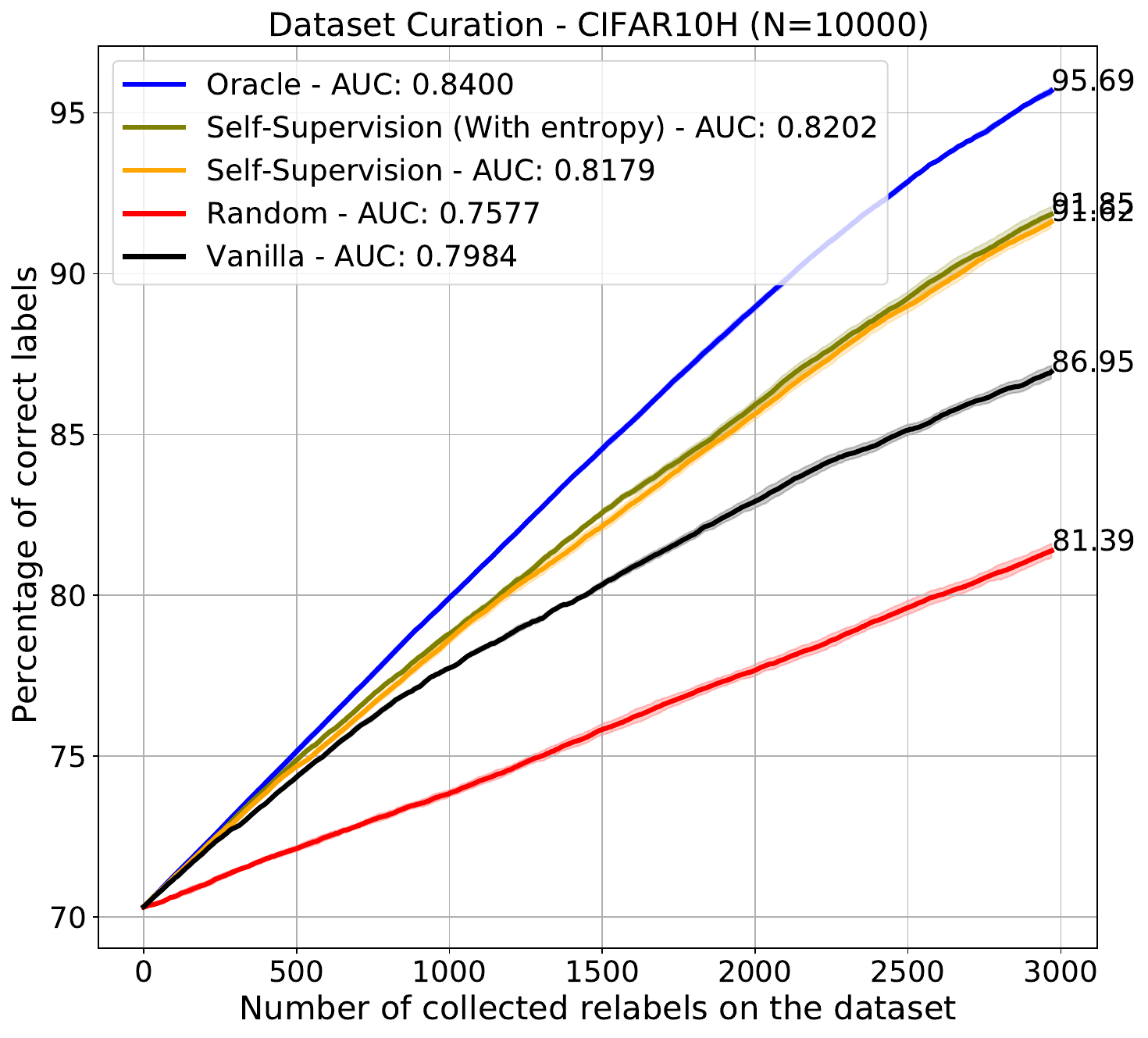}
        \includegraphics[width=0.3\linewidth]{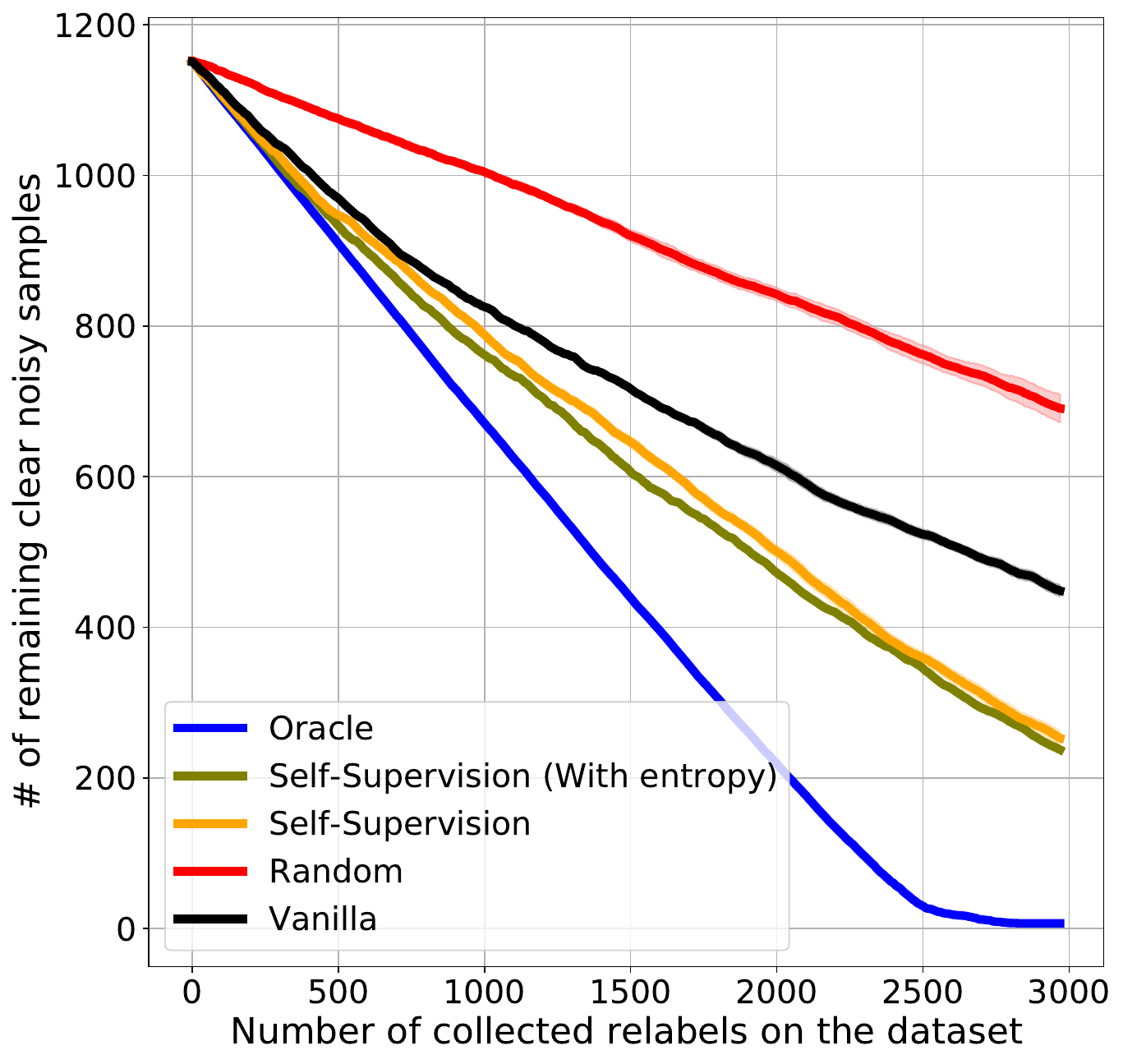}
        \includegraphics[width=0.3\linewidth]{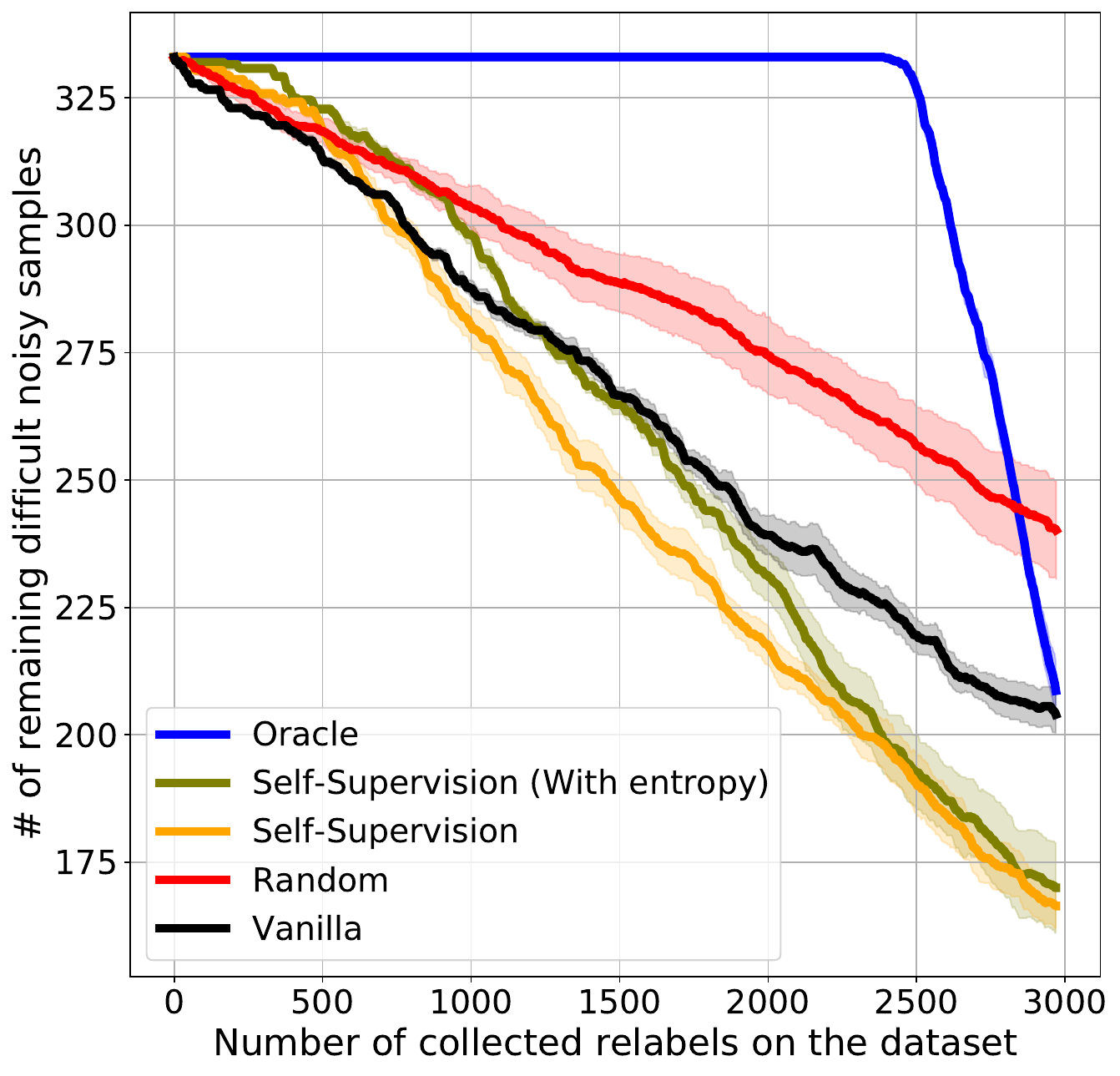}
        \caption{CIFAR10H -- Noise Rate 30\%}
    \end{subfigure}%
    \vspace{5 mm}
    \begin{subfigure}{\linewidth}
        \centering
        \includegraphics[width=0.32\linewidth]{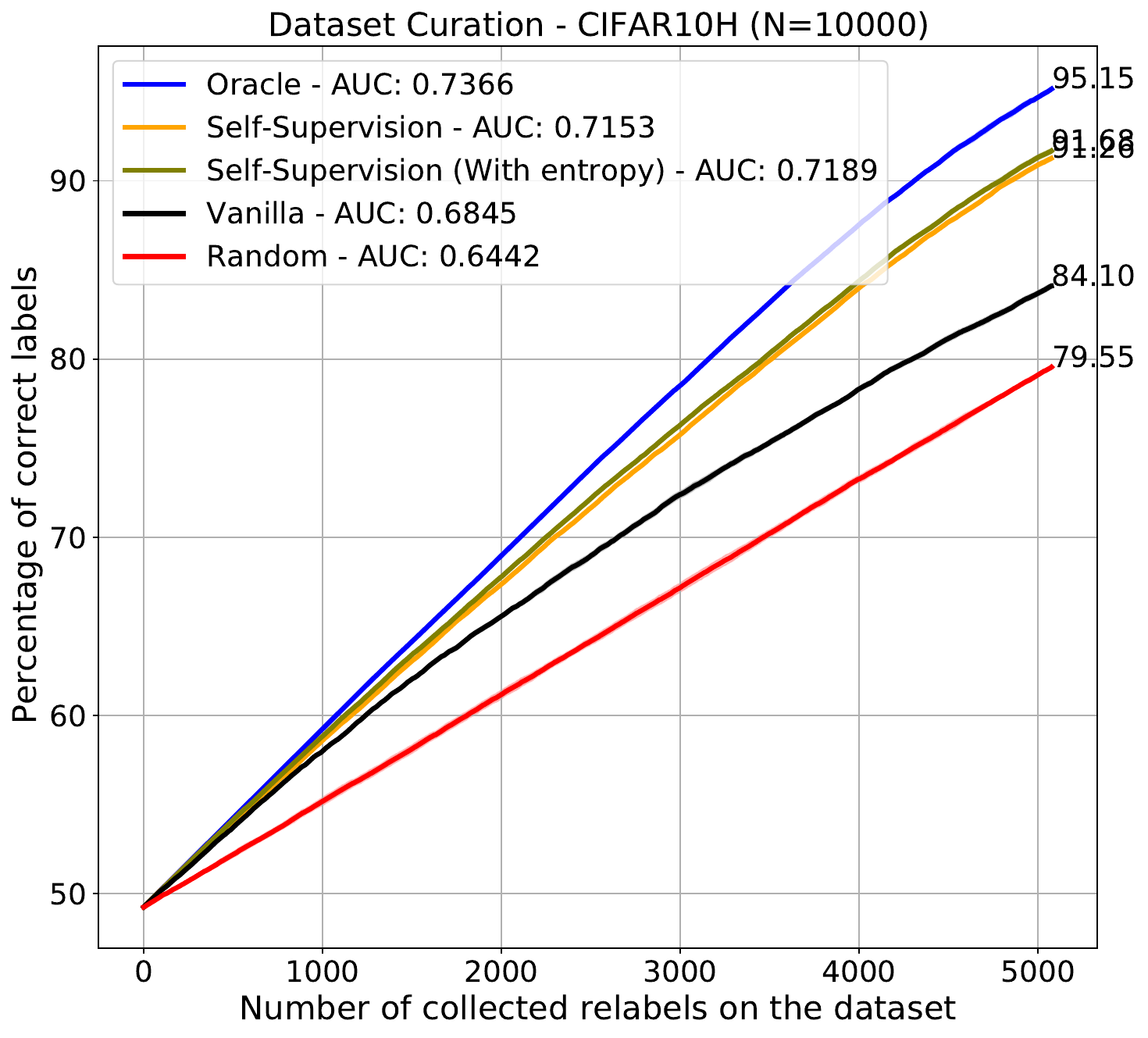}
        \includegraphics[width=0.3\linewidth]{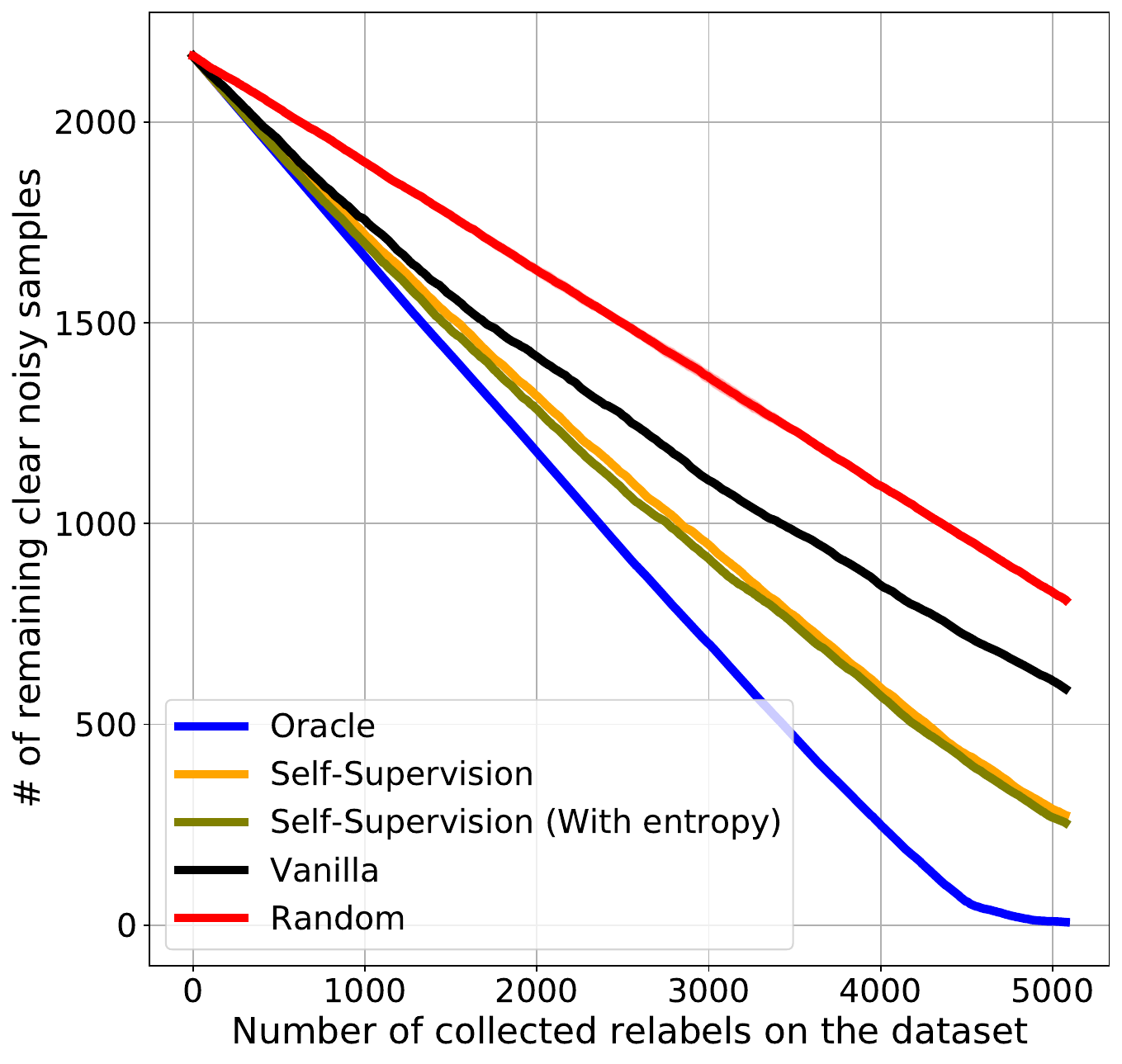}
        \includegraphics[width=0.3\linewidth]{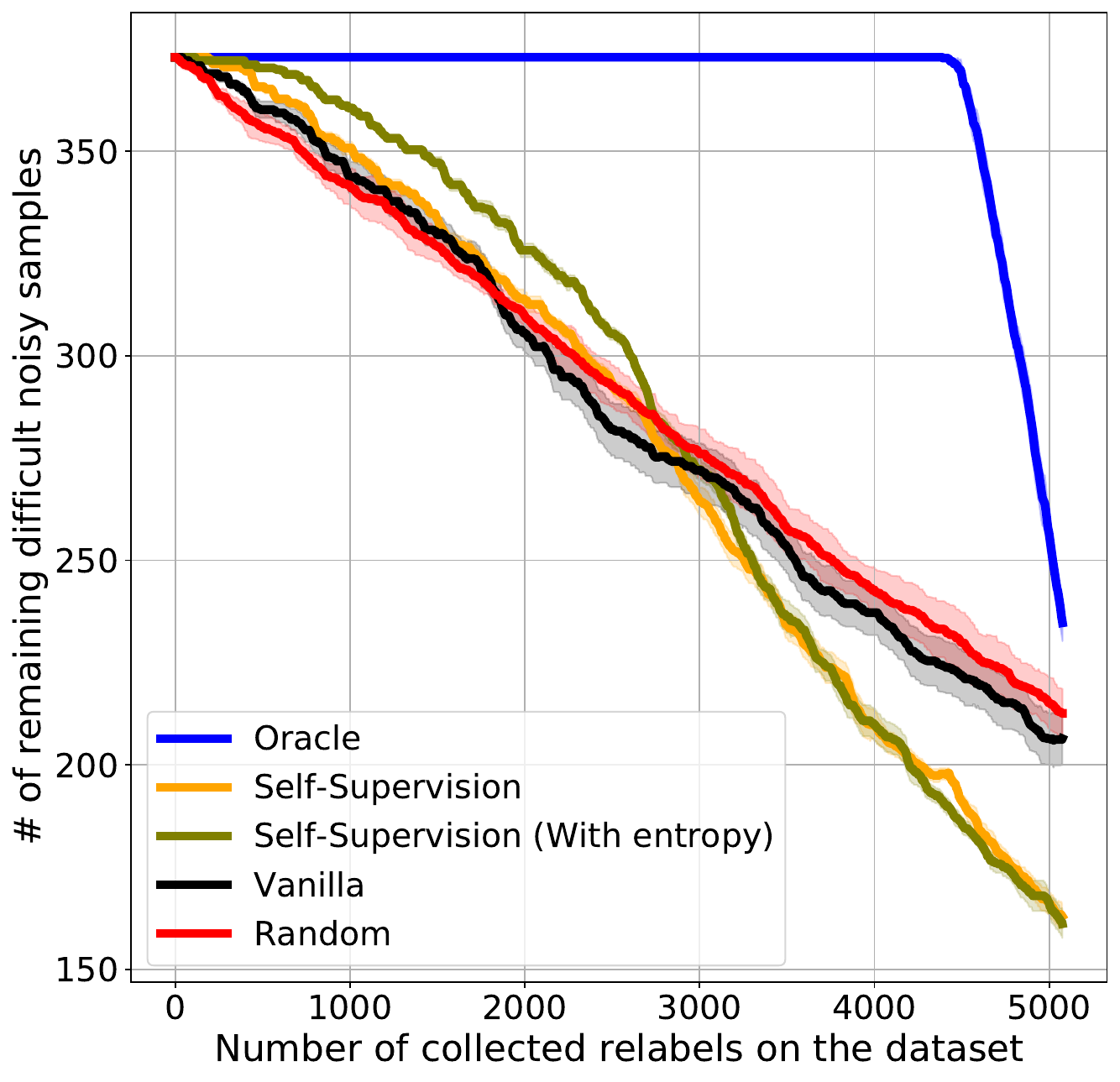}
        \caption{CIFAR10H -- Noise Rate 50\%}
    \end{subfigure}
    \caption{Results of noisy label cleaning on CIFAR10H dataset with different noise rates. The figures on the left illustrate the percentage of correct labels with respect to the number of relabels collected in the simulation. Here we see that by including the self-entropy term in \textit{SSL} selector (green), the final outcome can be slightly improved ($91.85$ vs $91.02$). Similarly, the number of remaining clear and difficult label noisy cases are shown in the middle and right, respectively. Best performing methods prioritise clear label noise cases over the ones with more labelling difficulty. Lastly, we see that this behaviour is further emphasised by including the ambiguity term in the sample scoring function (green vs orange curves). Shaded areas represent +/- standard deviation over 5 random seeds for relabelling.}
    \label{fig:sample_counts_CIFAR}
\end{figure*}

To better understand the impact of each term in the scoring function $\Phi$, we conducted simulation experiments by dropping the self-entropy term that captures the sample ambiguity. We hypothesise that this term encourages selectors to give a higher priority to clear label noise cases over the samples with higher labelling difficulty. To experimentally verify this, we first split the noisy samples into two disjoint subsets: clear noisy and difficult noisy. This is done by thresholding the normalised entropy of each sample's true label distribution at 0.3. Then we tracked the size of both noisy sets throughout the active label cleaning procedure to understand the sample ranking pattern. The experiments are conducted on the CIFAR10H dataset with 30\% and 50\% noise rates. The corresponding results are shown in \cref{fig:sample_counts_CIFAR}. We observe that by including the self-entropy term, the \textit{SSL} selector first prioritises the clear label noise cases as highlighted by the difference between green and orange curves in the plot (middle and right), which also yields a slight performance improvement in terms of mislabel correction rate. Moreover, we observe that the best performing sample selectors favour correcting first the clear label noise cases as it can be seen by the difference between the \textit{SSL} and \textit{Vanilla} approaches.
\begin{figure}[h!]
    \centering
    \figstyle
    \includegraphics[width=\linewidth]{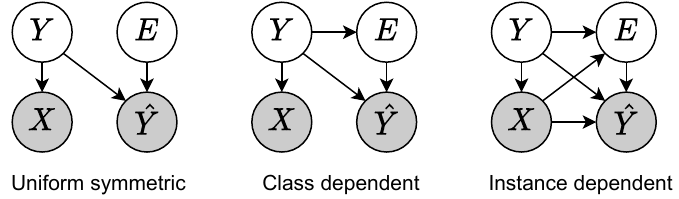}
    \caption{\textbf{Different label noise models used in robust learning.} The statistical dependence between input image ($X$), true label ($Y$), observed label ($\hat{Y}$), and error occurrence ($E$) is shown with arrows. Adapted from Fr\'enay et al.\cite{frenay2014classification}.}
    \label{fig:noise_model}
\end{figure}
\begin{figure*}[t!]
  \centering
  \begin{subfigure}{0.42\linewidth}
  \centering
  \figstyle
  \includegraphics[width=0.9\linewidth]{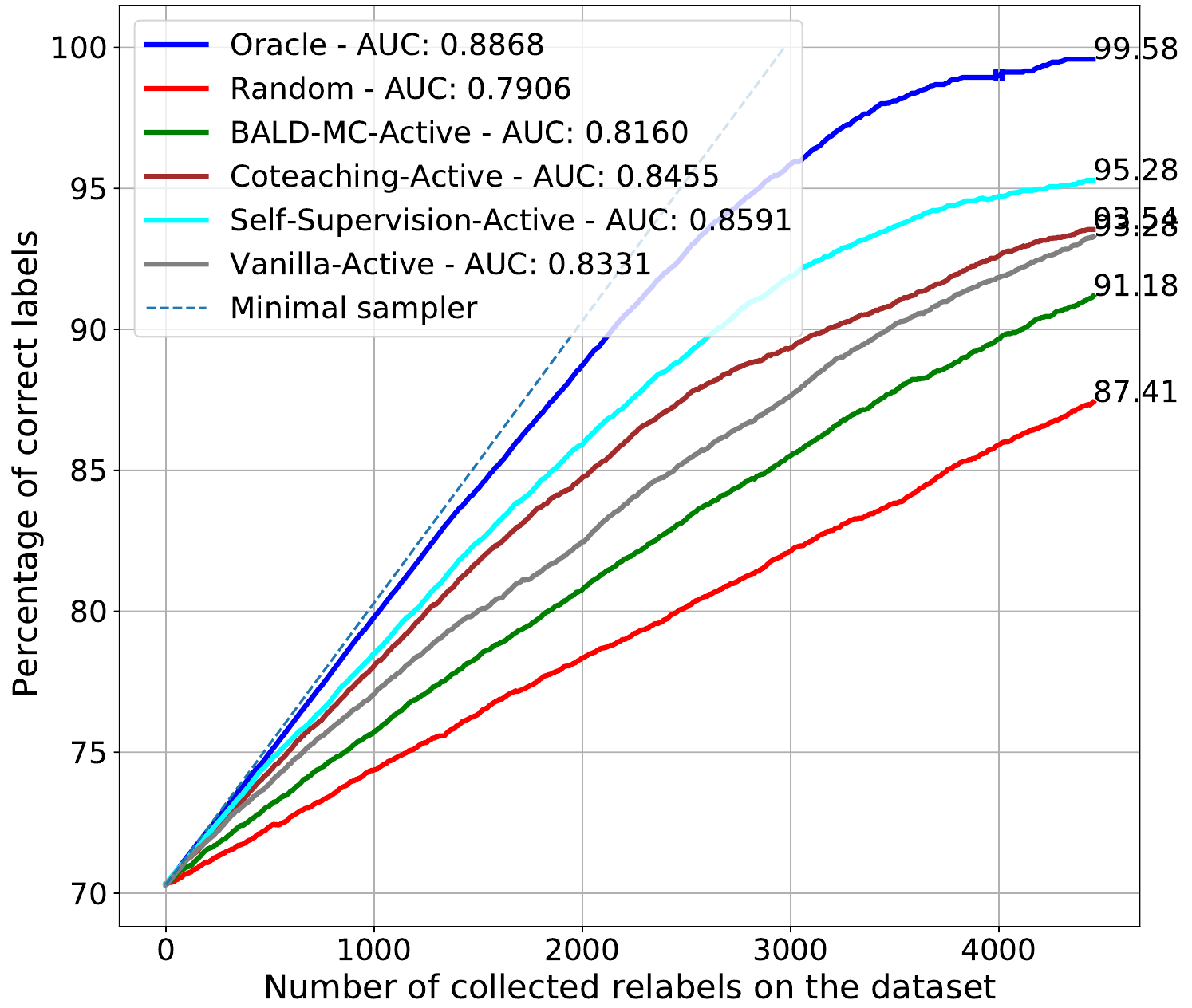}
  \label{fig:active_vs_our_result}
  \caption{CIFAR10H with $\eta=30\%$ initial noise rate}
  \end{subfigure}
  \begin{subfigure}{0.42\linewidth}
    \centering
    \figstyle
    \includegraphics[trim={1cm 0cm 0.3cm 1cm}, clip, width=0.9\linewidth]{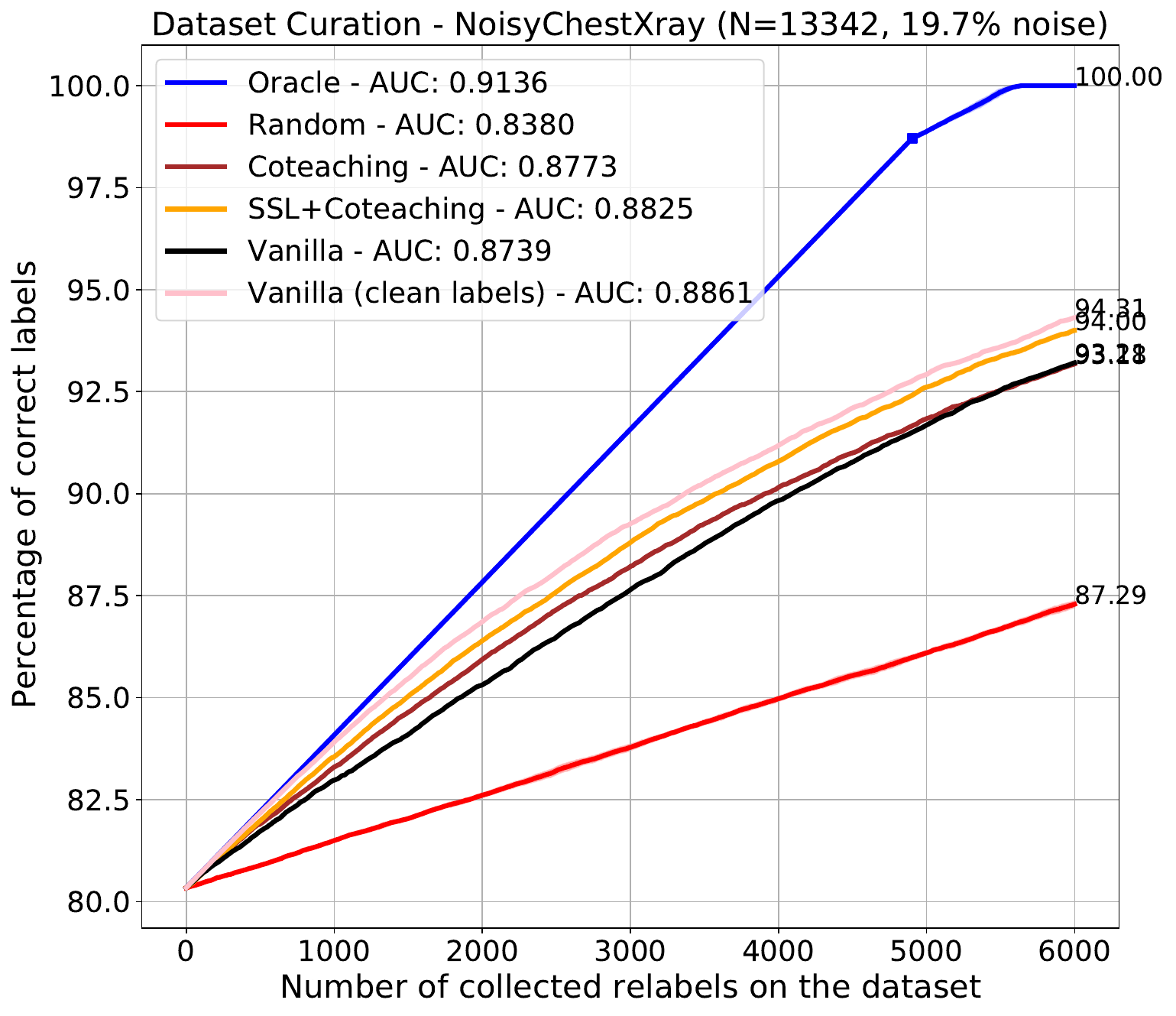}
    \caption{\NoisyCXR\ with $\eta=19.7\%$ initial noise rate}
    \label{fig:sim_chest_20}
  \end{subfigure}
  \caption{Correctness of labels (y-axis) with respect to re-labelling budget (x-axis) for higher noise rate for CIFAR10H and \NoisyCXR. Fig (a) shows that standard active learning based sample scoring functions, such as BALD \cite{gal2017deep} (in green), do not necessarily prioritise noisy labels in the relabelling procedure. Shaded areas represent +/- standard deviation over 5 random seeds for relabelling.}
\end{figure*}

\begin{figure}
    \centering \figstyle
    \includegraphics[width=0.65\linewidth]{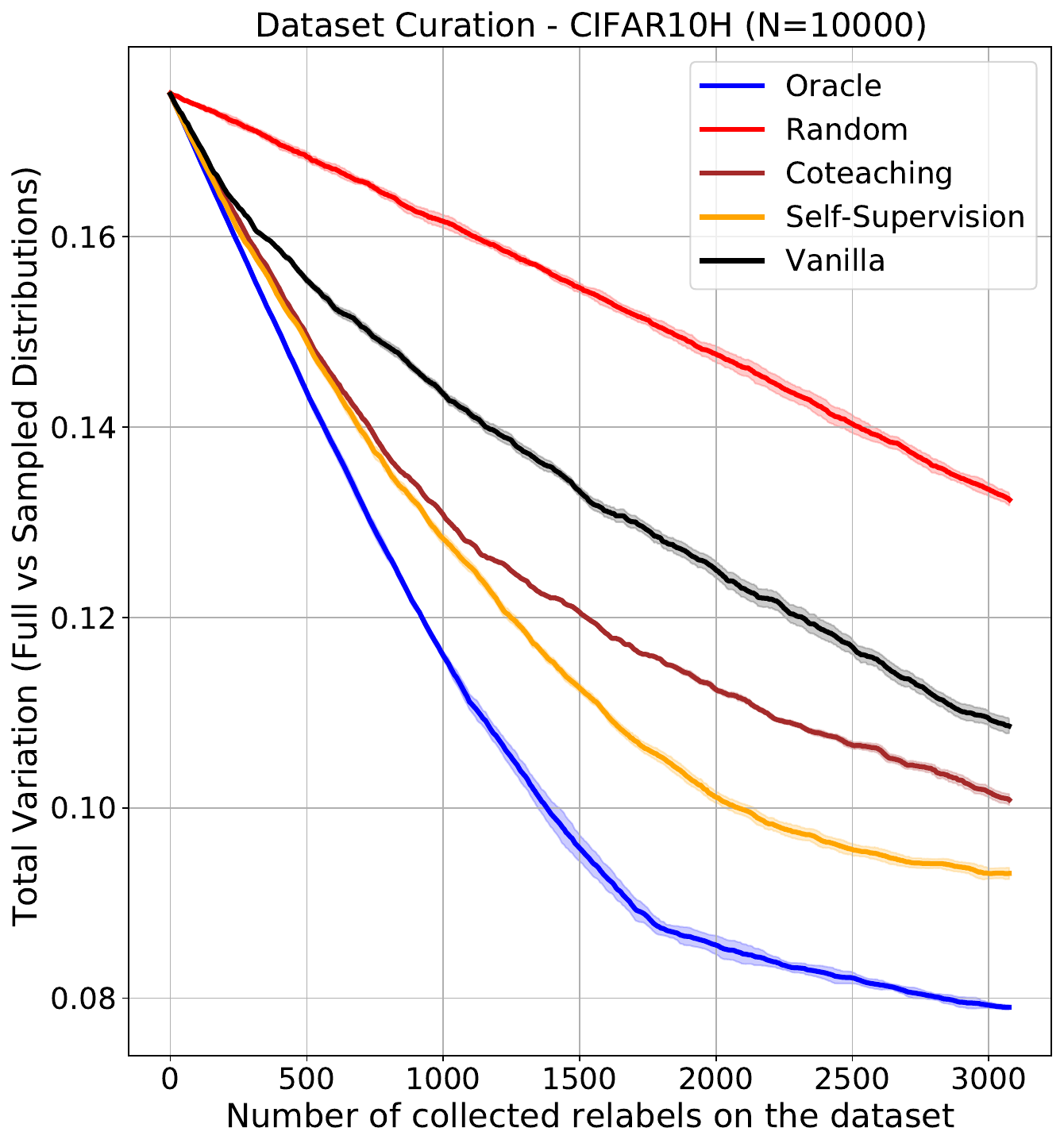}
    \caption{Total variation (y-axis) between true label distribution and distribution of sampled labels in function of relabelling budget (x-axis). A similar performance ranking is observed as in the experiments measuring the accuracy of majority labels (CIFAR10H, $\eta=$ 30\%). Shaded areas represent +/- standard deviation over 5 random seeds for relabelling.}
    \label{fig:total_variation}
\end{figure}

\begin{figure}
    \centering \figstyle
    \includegraphics[width=0.80\linewidth]{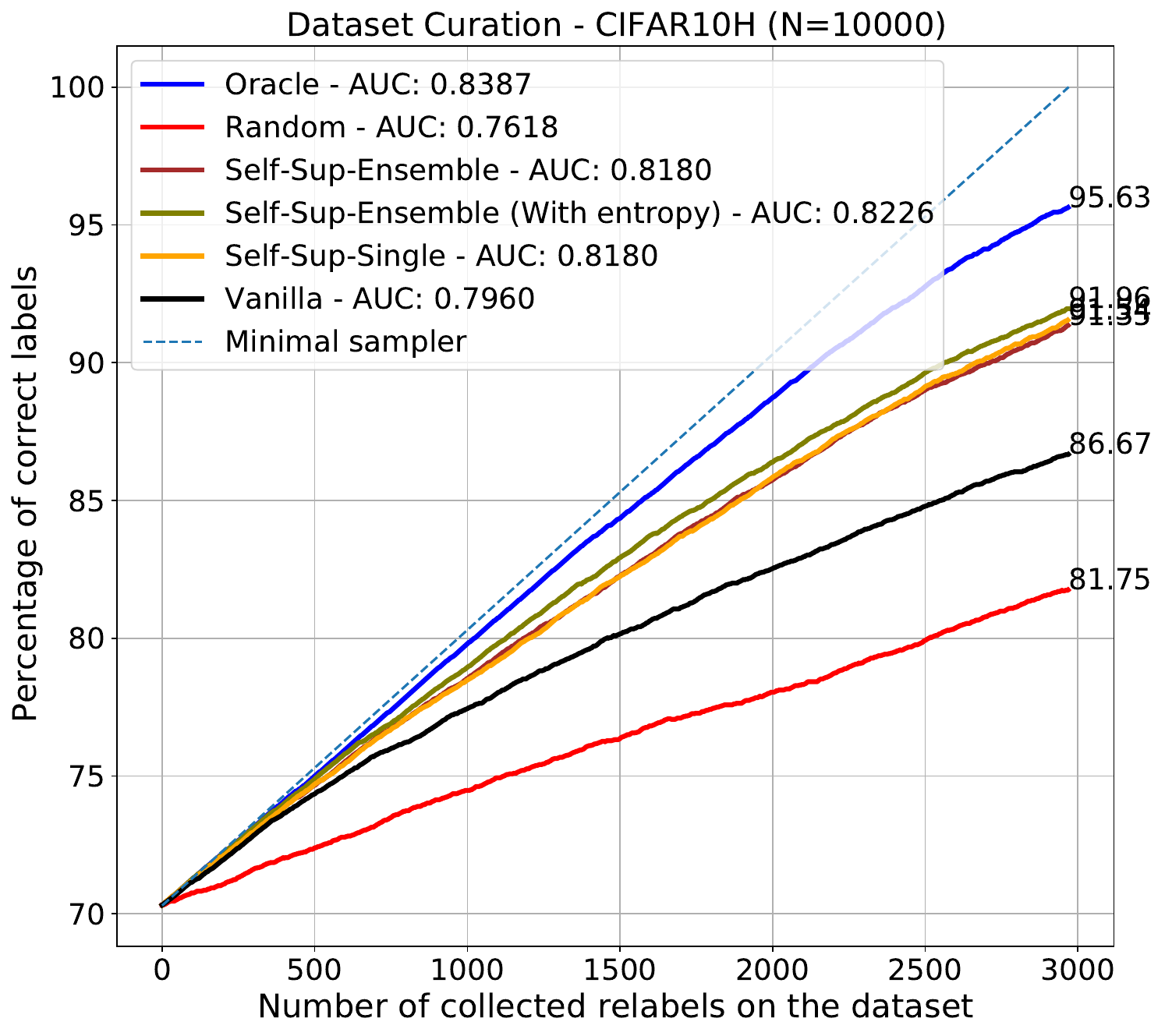}
    \caption{Simulation experiment benchmarking sample selectors that are based on single SSL model and ensemble of them. Uncertainty estimates become more reliable (green curve) when the entropy term is marginalised over model weights (CIFAR10H dataset, $\eta=$ 30\%). Shaded areas represent +/- standard deviation over 5 random seeds for relabelling.}
    \label{fig:ensemble_entropy_simulation}
\end{figure}

\subsection{Model ensembles yield better aleatoric uncertainty estimates.}
The ambiguity term in the scoring function ($\Phi$) can be better estimated when the posterior entropy is marginalised over a weight distribution as in the case of BALD \cite{houlsby2011bald}. To verify this hypothesis, an ensemble of SSL models are formed by training with 5 different seeds of weight initialisation whilst using the same set of initial labels. The simulations are repeated 5 times with different label sampling patterns. Supp. Table \ref{tab:ensemble_results} provides the AUC values for these experiments, where we see that the ambiguity term (Ensemble-Ent) improves the results further when ensembles are used and averaged posteriors (Ensemble) do not yield drastic improvement. \cref{fig:ensemble_entropy_simulation} shows this comparison for different resource constraint ($B$) values.

\begin{table}[h!]
    \tablestyle
	\caption{Impact of model ensembles on label cleaning performance and uncertainty estimation}
	
	{\centering
	\label{tab:ensemble_results}
	\setlength{\tabcolsep}{3pt}
	\begin{tabular}{@{\extracolsep{1pt}}llcccc@{}}
	    \toprule
		&& \multicolumn{4}{c}{(AUC) Perc. of correct labels vs number of relabels}\\
		\cmidrule(lr){3-6}
		&Model & Single & Single-Ent & Ensemble & Ensemble-Ent \\
		\midrule
		& SSL  & .818  & .819  & .820  & .823 \\
		\bottomrule
	\end{tabular} \\[1.5 mm]
	}
	\scriptsize
    The simulations are run over 5 different seeds of annotation sampling procedure. To form an ensemble, 5 different models are trained with different weight initialisations whilst keeping the initial labels the same (CIFAR10H, $\eta=$ 30\%, $B=3k$).
    
\end{table}

\subsection{Impact of model update frequency:} An ablation study is conducted to understand the influence of updating model's beliefs during the relabelling process (see \cref{tab:active_parameter_update}). In that regard, experiments are repeated for different $b$ values which determines the frequency of model updates using the collected labels. With these updates, an improved performance is observed on all model based approaches, which converge towards the upper-bound set by a model trained with all clean labels.

\begin{table}[t]
    \tablestyle
	\caption{Ablation study on the frequency of model updates ($b$) in active relabelling procedure.}
	\label{tab:active_parameter_update}
	\setlength{\tabcolsep}{3pt}
	\begin{tabular}{@{\extracolsep{1pt}}llcccc@{}}
	    \toprule
		&& \multicolumn{4}{c}{\% of Correct Labels (std) and AUC}\\
		\cmidrule(lr){3-6}
		&Model & $b=500$ & $b=1000$ & $b=2000$ & $b=\infty$ \\
		\midrule
		{\scriptsize\multirow{5}{*}{\rotatebox[origin=c]{90}{\textit{\NoisyCXR}}}}&
		Oracle  & -    & -    & -    & 100.0 (0), .946 \\
        &\multirow{2}{*}{Vanilla} & 94.81 (.08) & 94.82 (.04) & 94.75 (.05) & 94.34 (.03) \\
		&                     & .916 & .915 & .914 & .913 \\
		        &\multirow{2}{*}{SSL} & 95.12 (.04) & 95.04 (.04) & 95.02 (.04) & 95.03 (.02) \\
		&                     & .919 & .919 & .918 & .918 \\
		\midrule
		{\scriptsize\multirow{5}{*}{\rotatebox[origin=c]{90}{\textit{CIFAR10H}}}}&
		Oracle  & -    & -    & -    & 99.38 (.19), .887\\
		&\multirow{2}{*}{Vanilla} & 93.23 (.11) & 93.14 (.25) & 92.72 (.36) & 91.27 (.34)\\
		&                     & .836 & .836 & .834 & .829 \\
        &\multirow{2}{*}{SSL} & 95.21 (.20) & 95.12 (.19) & 94.97 (.19) & 94.94 (.17) \\
		&                     & .861 & .860 & .858 & .858 \\
		\bottomrule
	\end{tabular} \\[1.5 mm]
	\scriptsize
	 Model update frequency impact on the number of corrected labels is assessed on CIFAR10H (30\% initial noise,  $B=4.5k$) and \NoisyCXR\ (12.7\% initial noise, $B=4k$) for \textit{vanilla} and \textit{SSL-Linear}. The simulations are run over 5 different seeds of model fine-tuning and label sampling.
\end{table}

\subsection{Instance-dependent noise model.}
\label{sec:noise_models}
Uniform and class-dependent noise models, illustrated in \cref{fig:noise_model}, do not have a direct statistical dependency on the image content. This may lead researchers to design and experiment with artificial noise models that may not always hold true in practical scenarios. For instance, it would be more difficult to label low-resolution images of brown horses in comparison to white ones, since class confusion probability between horse and deer is likely to be higher. 

To model such dependencies, recent work has proposed the use of instance-dependent noise (IDN) models \cite{chen2020beyond, xia2020part}. In our study, we re-implemented the IDN model proposed in \cite{xia2020part} (Algorithm 2), where instance-specific flip rates $\*p_i \in \Reals^C$ are computed for each instance $i$ using a set of weights $\{\*W_1, \*W_2, \ldots, \*W_C\}$ as $\*p_i = \*W_{y_i} \operatorname{vec}(\*x_i)$. The weights $\*W_c \in \Reals^{C \times L}$ are drawn from a normal distribution, $\operatorname{vec}(\*W_c) \sim \mathcal{N}(\*0, \*I)$. The individual class flip probabilities are later softmax-normalised across all classes and scaled with the desired noise rate to generate instance-specific class confusion matrices. In contrast to \cite{xia2020part}, here we utilise image embeddings generated with a pretrained image model (ResNet-50, ImageNet) instead of raw pixel values $\*x_i$. This is mainly intended to associate the generated class transition models with higher-level image semantics.

\section{Supplementary Note 2: Implementation Details}
\label{sec:implementation_details}

\subsection{CIFAR10H.}

\paragraph{Supervised models:} For all experiments, the classifier and selector are trained with the same model configuration but they differed in terms of the labels used to train them. For all experiments, we trained a ResNet-50 model with a bottleneck unit containing convolution layers with batch-normalisation. The models are trained for $160$ epochs with the SGD optimiser and Nesterov momentum ($0.9$) using a base learning rate of $0.1$ and weight decay of $10^{-4}$. The learning rate is reduced by a factor of $0.1$ at epochs $80$ and $120$. Each gradient update is computed over a mini-batch of $256$ input images by computing cross-entropy with respect to smoothed target labels ($\epsilon=0.1$). Images are augmented prior to model forward pass by applying standard natural image transformations (e.g.\ colour jitter, affine transformation, and horizontal flips). 

For the training of co-teaching models \cite{han2018co}, the drop-rate is set to match the expected noise rate in the datasets, which can be further tuned on a separate validation set for potentially improved performance. In the initial warm-up phase (up to epoch $10$), all samples are used in SGD updates, including the ones with large loss value.

\paragraph{SSL training:} For the training of BYOL \cite{grill2020bootstrap} self-supervision models, CIFAR10H dataset ($10k$) is used as a training set. For each true image pair, the standard image augmentations are applied, which are outlined in detail in the SimCLR study \cite{chen_simclr}. The encoder models are trained with Adam \cite{kingma2014adam} optimiser for $2500$ epochs with the base learning rate of $10^{-3}$ and batch size of $512$. As a learning rate scheduler, a cosine decay function is utilised. More importantly, in order not to bias the model comparison, the same ResNet-50 architecture is used as the base encoder for the BYOL models.

To build an image classifier on top of \textit{SSL} embeddings, a single linear layer is trained for $120$ epochs using both weight decay ($10^{-3}$) and tanh logit regularisation terms ($\alpha=20$, defined in Methods) in order to avoid over-fitting on the noisy labels. Similarly to the \NoisyCXR\ experiments, co-teaching could potentially be utilised together with \textit{SSL} weights without freezing the encoder parameters, which is omitted in CIFAR10H experiments as the linear head experimentally proved to be sufficient in identifying noisy labels.

\subsection{\NoisyCXR.}
\paragraph{Supervised models:} As for CIFAR10H, the classifier and selector are trained with the same model configuration but they differed in terms of the labels used to train them. For all experiments, a ResNet-50 model is trained with the Adam optimiser \cite{kingma2014adam} with cross-entropy loss, using a learning rate of $10^{-5}$ for models trained from scratch (resp.\ $10^{-6}$ for fine-tuning), weight decay of $10^{-4}$ and batch size 32. For all models, we use 50\% of the data for training and 50\% for validation. At training time, the under-represented class (``Pneumonia-like opacity") is over-sampled to handle the class-imbalance according to the class distribution observed in the training set. In terms of preprocessing, images are first augmented in native resolution (if applicable), then resized to $256 \times 256$, and finally they are centre-cropped to $224 \times 224$. At training time, the following augmentations are used: random horizontal flipping ($p=0.5$), random rotation ($[-30^{\circ}, 30^{\circ}]$), random shear transform ($[-15, 15]$), random contrast, random brightness, and random crop ($[80\%, 100\%]$). Models trained from random initialisation are trained for 150 epochs; models initialised from a pre-trained checkpoint are fine-tuned for 100 epochs. For ``SSL-Linear", we built a linear image classifier on top of the frozen SSL encoder, trained for 100 epochs with a starting learning rate of $10^{-4}$ decreased to $10^{-5}$ after 10 epochs; we used tanh logits regularization ($\alpha=10$) and reduced the set of augmentations to horizontal flips. 

For the training of co-teaching models \cite{han2018co}, the drop-rate is set to match the expected noise rate in the datasets. The warm-up phase was set to 20 epochs during which no samples are dropped (resp. 10 epochs when initialised from SSL checkpoint).

\begin{table}[t]
    \tablestyle
    \caption{The impact of image augmentations on the quality of learnt BYOL embeddings, when applied to medical images.}
    {
    \centering
    \label{tab:ssl_augment}
    \begin{tabular}{lc}
    \toprule
    Augmentations     &  ROC-AUC \\
    \midrule
    All w/out cropping & 0.853 \\
    All w/out brightness and contrast & 0.866 \\
    All w/out rotation and shear & 0.868 \\
    \midrule
    All augmentations (baseline)    & 0.871 \\
    \bottomrule
    \end{tabular} \\[3 mm]
    }
    \scriptsize
    The embedding quality is assessed in terms of linear separation of classes by training a linear-head on top of BYOL encoder output. At each experiment, one data augmentation is omitted to measure its impact, and classification results are obtained on the validation set using ROC-AUC.
\end{table}

\paragraph{SSL training:} The self-supervised models have been trained with BYOL \cite{grill2020bootstrap} using a ResNet-50 encoder. The final model is trained using the NIH dataset \cite{NIHDataset} (80\% training, 20\% validation), using an effective batch size of 4800 (batches of 600 pairs of images of size $224 \times 224$ on 8 GPUs) for 1000 epochs. The momentum parameter $\tau$ used in BYOL teacher encoder was set to $0.99$. As augmentations, we use random horizontal flipping (with probability $p=0.5$), random rotation and shear (drawn from $[-180^{\circ}, 180^{\circ}]$ and $[-40,40]$), random crop ($[40\%, 100\%]$), CutOut (masking out rectangles of varying size, size covering 15--40\% of the image), elastic transforms \cite{simard2003best} ($a=34$, $s=4$, applied with probability $0.5$), random contrast and brightness changes, and adding Gaussian noise ($\sigma=0.01$, applied with probability 0.5).

We ran experiments on the \NoisyCXR\ dataset to determine the impact of each of those augmentations on the quality of the resulting embedding. To this end, a linear classifier is learnt on top of image embeddings during BYOL training and its performance is monitored with ROC-AUC metric on the validation set. By disabling one augmentation at a time and comparing to the baseline with all augmentations, we see that image cropping contributes most to the results, followed by contrast and brightness augmentations, then rotation and shear, results are provided in \cref{tab:ssl_augment}. Using the full set of augmentations often meant that performance kept increasing when training for longer, whereas using fewer augmentations led to performance plateauing earlier in training. Histogram normalisation had a noticeable impact on the result. Using the native images without histogram normalisation led to an AUC roughly 0.6\% higher than with histogram normalisation.

\footnotesize
\bibliographystyle{naturemag}
\bibliography{references}